\documentclass[sigconf]{acmart} 

\usepackage{amsthm}
\usepackage{framed}
\usepackage{graphicx}
\usepackage{multirow}
\usepackage{subcaption} 
\usepackage{wrapfig}
\usepackage{xcolor}
\usepackage{natbib}
\usepackage[ruled,vlined]{algorithm2e}
\usepackage[figuresright]{rotating}
\usepackage{algorithmic}

\AtBeginDocument{%
  }

\begin{document}

\title{Consistent Point Orientation for Manifold Surfaces via Boundary Integration  }



\author{Weizhou Liu}
\email{liuweizhou@mail.bnu.edu.cn}
\orcid{0009-0007-9127-8313}
\affiliation{
\institution{Beijing Normal University}
\country{China}
}

\author{Xingce Wang}
\authornote{Co-corresponding authors: Xingce Wang and Ying He.}
\email{wangxingce@bnu.edu.cn}
\orcid{0000-0002-3177-8902}
\affiliation{
\institution{Beijing Normal University}
\country{China}
}

\author{Haichuan Zhao}
\email{201931210007@mail.bnu.edu.cn}
\orcid{0000-0003-4967-7533}
\affiliation{
\institution{Beijing Normal University}
\country{China}
}

\author{Xingfei Xue}
\email{xuexf@mail.bnu.edu.cn}
\orcid{0009-0003-9411-5350}
\affiliation{
\institution{Beijing Normal University}
\country{China}
}

\author{Zhongke Wu}
\email{zwu@bnu.edu.cn}
\orcid{0000-0003-3735-6476}
\affiliation{
\institution{Beijing Normal University}
\country{China}
}

\author{Xuequan Lu}
\email{b.lu@latrobe.edu.au}
\orcid{0000-0003-0959-408X}
\affiliation{
\institution{La Trobe University}
\country{Australia}
}

\author{Ying He}
\authornotemark[1]
\email{yhe@ntu.edu.sg}
\orcid{0000-0002-6749-4485}
\affiliation{
\institution{Nanyang Technological University}
\country{Singapore}
}

\begin{abstract}
This paper introduces a new approach for generating globally consistent normals for point clouds sampled from manifold surfaces. Given that the generalized winding number (GWN) field generated by a point cloud with globally consistent normals is a solution to a PDE with jump boundary conditions and possesses harmonic properties, and the Dirichlet energy of the GWN field can be defined as an integral over the boundary surface, we formulate a boundary energy derived from the Dirichlet energy of the GWN. Taking as input a point cloud with randomly oriented normals, we optimize this energy to restore the global harmonicity of the GWN field, thereby recovering the globally consistent normals. Experiments show that our method outperforms state-of-the-art approaches, exhibiting enhanced robustness to noise, outliers, complex topologies, and thin structures.
Our code can be found at \url{https://github.com/liuweizhou319/BIM}.
\end{abstract}

\maketitle

\section{Introduction}
Point clouds with globally consistent normals have numerous downstream applications, such as point set upsampling~\cite{EAR2013, feng2022np}, point cloud filtering~\cite{zhang2020pointfilter,DBLP:journals/tvcg/LuSLMH22}, surface reconstruction~\cite{kazhdan2006poisson,huanghui2009,SSD, kazhdan2013screened, gauss_R, Occupancy_Networks, SPSR2022, alec2023NeuralPoisson}, and segmentation~\cite{KHALOO20171}.
As a fundamental problem in computer graphics, it remains challenging due to the complexity of object geometry, topology, and the influence of noise. 

Most existing methods~\cite{hoppe1992surface,SGP2003, Pau2003, konig2009consistent, Towards2017,metzer2021orienting} estimate normals perpendicular to the point cloud surface and then use propagation to flip inconsistent normals. As pointed out by \citet{li2023neural}, propagation strategies are affected by the direction of distribution of the unoriented normal vectors and often lead to undesirable results. Moreover, neighborhood-based normal estimation methods tend to perform poorly when dealing with noise and geometries with small reach. In the past few years, there has been significant advancement in the field of globally consistent normal estimation. iPSR \cite{hou2022iterative} reconstructs surfaces from unoriented point clouds by iteratively refining the normals obtained from the PSR solver~\cite{kazhdan2006poisson}. While effective for dense point clouds, iPSR encounters difficulties with sparse inputs. In sparsely sampled regions, the potential generation of erroneous normals by iPSR can result in disconnected components in the reconstructed surface. There are also deep learning-based methods for point normal estimation or orientation~\cite{PCPNET2018,Ben_Shabat_2019_CVPR, zhu2021adafit,li2023neural,li2023neuralgf, li2023shsnet,li2023NeAF}.
While these methods enhance the precision of predictions, they typically do not reverse the normals when the prediction of normal orientation is inconsistent.

The generalized winding number (GWN)~\cite{jacobson2013winding} has proven to be a powerful tool for robust segmentation of 3D surfaces into distinct inside and outside regions. Its application extends to the orientation of triangle meshes and point clouds, showcasing its versatility in the field of geometric processing. Building on this foundational work, \citet{TakayamaJKS14} proposed to orient triangle soups by minimizing the Dirichlet energy associated with the GWN field. \citet{Barill2018FastWN} developed a point cloud-based approach to estimating the GWN, linking point cloud normals to the generalized winding number. More recently, PGR \cite{lin2022surface} and GCNO \cite{xu2023globally} adopted point-based GWN for orienting point clouds. They regularize the values of the GWN both inside and outside the point cloud. Despite these advancements, these methods are usually limited to small-scale models due to their high computational cost. Additionally, they are not effective in handling models with complex topology and often fail to robustly address issues posed by thin structures and noise.

Our work aims at developing a robust approach for computing globally consistent normal orientations in point clouds sampled from manifold surfaces. We identify harmonicity and jump boundary conditions as two key factors in point cloud orientation. Leveraging the fact that the Dirichlet energy of the GWN field can be expressed as an integral over the boundary~\cite{TakayamaJKS14}, we introduce a new boundary energy formulation. In this framework, the normals associated with each point are variables. This boundary energy coincides with the Dirichlet energy only when the normals are globally consistent. By imposing constraints on the range of the GWN values, we observe an inverse correlation between boundary energy and Dirichlet energy: the boundary energy is small for randomly oriented normals, and becomes large when normals are globally consistent. This observation motivates us to maximize the boundary energy from initially random normals. Employing an efficient L-BFGS solver, we iteratively enhance the global harmonicity of the generalized winding number field, eventually yielding globally consistent normals. Extensive experiments demonstrate the effectiveness of our method, achieving high-quality orientation and reconstruction results. 

\section{Related Work}

The problem of global point orientation dates back to the 1990s, in the seminal work of \citet{hoppe1992surface}. The authors calculated initial normals utilizing principal component analysis, and subsequently corrected inconsistent normals through propagation. This pioneering work has inspired many follow-up works, including \cite{konig2009consistent, jakob2019parallel,levin1998approximation,mitra2003estimating,alliez2007voronoi,CGF2012, Rui2022RFEPS}, among others. Although these methods perform well with simple models, their performance deteriorates in the presence of complex geometries and data imperfections, especially in the presence of noise and outliers. 

Reconstructing 3D surfaces from unoriented point clouds is closely related to the problem of point orientation. A popular approach for 3D reconstruction involves using an implicit function to represent the target surface, where the gradients of this function at a particular level set serve as the point normals. Current approaches predict globally consistent normals by using signed/unsigned distance fields or by considering point clouds as dipoles to generate generalized winding number fields. \citet{mullen2010signing} computed an unsigned distance field for the input point cloud, subsequently determining the sign by minimizing the Dirichlet energy. NGLO  \cite{li2023neural} enhanced normal accuracy by optimizing gradients based on the fitted distance field. NeuralGF \cite{li2023neuralgf} introduced a novel paradigm of learning gradients for fitting a signed distance field with neural implicit functions, enabling unsupervised point cloud normal estimation. The Dipole method \cite{metzer2021orienting} maximizes the electric field gradient of the point cloud using a greedy strategy, aiming to approximate the generalized winding number. iPSR \cite{hou2022iterative} iteratively refined the surface by feeding the normals obtained from the previous iteration into the PSR \cite{kazhdan2013screened} solver, gradually improving the surface quality. PGR \cite{lin2022surface} treats normals and surface element areas as unknown parameters, interpreting the indicator function as a member of a parametric function space using the Gauss formula, and recovering the normals by estimating the values of the indicator function. GCNO \cite{xu2023globally} incorporated the prior knowledge of the generalized winding number field as an approximation to the indicator function for point clouds with globally consistent normals. Despite significant advancements in the field of 3D reconstruction, existing methods for reconstructing 3D surfaces from unoriented point clouds still suffer from various issues. For example, NGLO's dependency on a training set limits its ability to robustly handle diverse types of data and noise levels. Dipole may generate conflicting normals at thin structures. iPSR can lead to disconnected components in cases of sparse point clouds. PGR lacks robustness to noise and demands high computational resources. GCNO's effectiveness is contingent upon an even distribution of Voronoi vertices both inside and outside the point cloud--a condition that, when unmet, significantly degrades the quality of the results.

PDEs with jump boundary conditions are widely used in surface reconstruction, rendering, and fluid simulation. In surface reconstruction, jump boundary conditions are used to infer the implicit function from the point cloud with consistent normals. The indicator function obtained from the PSR \cite{kazhdan2006poisson, kazhdan2013screened} solver effectively satisfies the jump boundary conditions. The generalized winding number generated by point clouds with globally consistent normals \cite{Barill2018FastWN} can be regarded as an approximate solution to the Poisson equation with jump boundary conditions. The surface can be reconstructed by extracting the iso-surfaces from the generalized winding number. In rendering, Poisson vector graphics \cite{Hou2020PoissonVG} and diffusion curves \cite{orzan2008diffusion} treat the input strokes as jump boundary conditions and accomplish color rendering by solving Laplace's equation. In fluid simulation \cite{Leal2007AdvancedTP}, jump boundary conditions are used to represent the physical properties at the interface between two fluids. Recently, \citet{Feng2023WindingNO} proposed a method for computing the winding number for discrete surfaces. They approached the problem by considering the relationship between the winding number and harmonic functions and solved it by solving Laplace's equation with jump boundary conditions.

\section{Preliminaries}
Given a solid object $\Omega\subset\mathbb{R}^3$, we denote its boundary surface by $\partial \Omega$. Let $\mathcal{P} = \{\mathbf{p}_i | \mathbf{p}_i \in \partial\Omega\}_{i=1}^{n}$ be a point cloud sampling the manifold surface. For each point $\mathbf{p}_i$, we denote by $\hat{\mathbf{n}}_i$ its outward unit normal. We denote the inner product in $\mathbb{R}^3$ by $\langle \cdot, \cdot \rangle$.

\subsection{Poisson's Equation with Jump Boundary Conditions}
Poisson's equation with jump boundary conditions finds extensive applications across physics, scientific computing, simulation, and geometric processing. The equation is formulated as:
\begin{equation}
    \Delta u(\mathbf{x}) = f(\mathbf{x}), \;\;\mathbf{x} \in\; \mathbb{R}^3 \setminus \partial \Omega \label{poissonEqu},    
    \end{equation}
subject to the Dirichlet boundary condition 
    \begin{equation}
    u^{+}(\mathbf{x}) - u^{-}(\mathbf{x}) = k, \;\; \mathbf{x} \in\; \partial \Omega,\label{eqn:jumpboundary}
    \end{equation} 
and the Neumann boundary condition
    \begin{equation}
    \partial u^{+}(\mathbf{x})/\partial \mathbf{n}_{\mathbf{x}} =  \partial u^{-}(\mathbf{x})/\partial \mathbf{n}_{\mathbf{x}}, \;\;\mathbf{x} \in\; \partial \Omega, \label{eqn:jumpboundary_Neumann}
\end{equation} 
where $u:\mathbb{R}^3 \to \mathbb{R}$ is the sought solution, $f:\mathbb{R}^3 \to \mathbb{R}$ represents a source term, and $k$ is a constant. The term $\mathbf{n}_{\mathbf{x}}$ denotes the globally consistent outward normal at $\mathbf{x}$. We also define exterior and interior boundary values as the limiting values of $u$ at the boundary from the exterior and interior, respectively. That is, $u^\pm(\mathbf{x}) = \lim_{\epsilon \to 0}u(\mathbf{x} \pm \epsilon \mathbf{n}_\mathbf{x})$ for $x \in \partial\Omega$. When $f=0$, the solution $u$ is harmonic, and in this context, $u$ also minimizes the Dirichlet energy functional $\int_{\mathbb{R}^3 \setminus \partial \Omega}\|\nabla u\|^2$.

\subsection{Boundary Integral Equation}
The boundary element method is a powerful tool for solving partial differential equations. BEM transforms the differential equation within the computational domain into a boundary integral equation. The boundary is then discretized into elements, allowing the discretization of the boundary integral equation as a dense system of linear equations without requiring explicit discretization of the computational domain.

 Equation~\eqref{poissonEqu} can be expressed in the form of a Boundary Integral Equation~\cite{COSTABEL1987243} using the Poisson kernel and Green's function as follows:
\begin{eqnarray}
&&u(\mathbf{x})=\int_{\mathbb{R}^3\setminus\partial\Omega}G(\mathbf{x}, \mathbf{y})f(\mathbf{y})\mathrm{d}\mathbf{y} \nonumber\\
    &+&\int_{\partial \Omega}P(\mathbf{x}, \mathbf{z})\left[u^{+}(\mathbf{z})-u^{-}(\mathbf{z})\right]
    -G(\mathbf{x}, \mathbf{z})\left[\frac{\partial u^{+}(\mathbf{z})}{\partial \mathbf{n}_\mathbf{z}}- \frac{\partial u^{-}(\mathbf{z})}{\partial \mathbf{n}_\mathbf{z}}\right]\mathrm{d}\mathbf{z}\nonumber\\
&=&\int_{\Omega}G(\mathbf{x}, \mathbf{y})f(\mathbf{y})\mathrm{d}\mathbf{y} +\int_{\partial \Omega}P(\mathbf{x}, \mathbf{z})\left[u^{+}(\mathbf{z})-u^{-}(\mathbf{z})\right]\mathrm{d}\mathbf{z}  
    \label{BIE}
\end{eqnarray}
where $G(\mathbf{x}, \mathbf{y})$ represents Green's function satisfying $\Delta G(\mathbf{x}, \mathbf{y})=\delta_\mathbf{y}(\mathbf{x})$. Here $\delta_\mathbf{y}(\mathbf{x})$ is Dirac delta function and $P(\mathbf{x}, \mathbf{y})$ denotes the Poisson kernel.
The final term in the second integral vanishes due to the Neumann boundary condition. 

\subsection{Generalized Winding Number}
The generalized winding number is the number of times a point in space is enclosed by a surface. As shown in~\cite{jacobson2013winding}, the generalized winding number $w$, generated by globally consistent outward normals, satisfies Laplace's equation. This condition represents a special case of 
Equation~\eqref{poissonEqu} when $k = -1$ and $f \equiv 0$. Consequently, we can express $w$ using Equation~\eqref{BIE} as 
\begin{equation}
w(\mathbf{x})=\int_{\partial \Omega}P(\mathbf{x}, \mathbf{z})d\mathbf{z}.
\label{GWN}
\end{equation}
The Poisson kernel $P(\mathbf{x}, \mathbf{y})$, which is the directional derivative of Green's function 
 in the direction $\mathbf{v}_\mathbf{y}$, is given by: 
$$
    P(\mathbf{x}, \mathbf{y})=
\frac{\partial G(\mathbf{x}, \mathbf{y})}{\partial \mathbf{v}_\mathbf{y}}
= \frac{1}{4\pi}\frac{\langle \mathbf{n}_\mathbf{y}, \mathbf{x}-\mathbf{y} \rangle}{\|\mathbf{x}-\mathbf{y}\|^3}.
$$
 
Given an oriented point cloud $\mathcal{P}$, where each point $\mathbf{p}_i$ is associated with a globally consistent outward normal $\mathbf{n}_i$, \citet{Barill2018FastWN} approximates the GWN for an arbitrary query point $\mathbf{q}\in\mathbb{R}^3$ by 
\begin{equation}w(\mathbf{q}):=\sum\limits_{i=1}^{n}{a_i}\frac{1}{4\pi}\frac{\langle \mathbf{n}_i, \mathbf{p}_i-\mathbf{q} \rangle}{\|\mathbf{p}_i-\mathbf{q}\|^3}
=\sum\limits_{i=1}^{n}a_i P(\mathbf{q}, \mathbf{p}_i).
    \label{GWN_calc}
\end{equation}
\begin{wrapfigure}[6]{r}{0.5\linewidth}
  \vspace{-12pt}
  \hspace{-20pt}
  \includegraphics[width=1.1\linewidth]{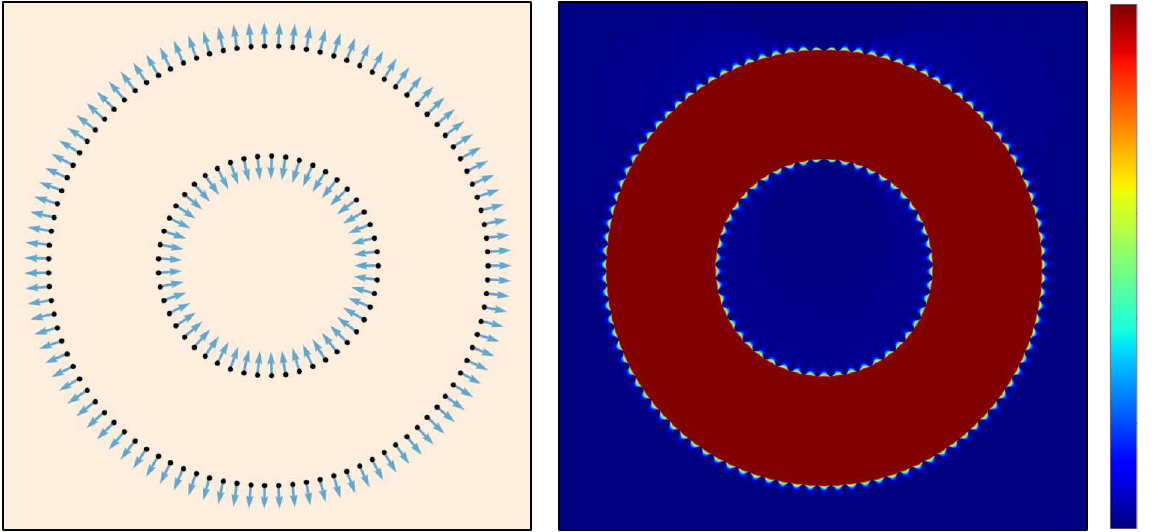}
  \vspace{-5pt}
  \put(-2,55){\makebox(0,0)[bl]{\small \textsf{1}}}
  \put(-2,0){\makebox(0,0)[bl]{\small \textsf{0}}}
  \label{fig:GWN}
\end{wrapfigure}
Here $a_i$ is the geodesic Voronoi area of point $\mathbf{p}_i$, representing the contribution of point $\mathbf{p}_i$ to the surface integral. Since computing geodesic Voronoi diagram on discrete surfaces is time consuming, \citet{Barill2018FastWN} suggested calculating $a_i$ as a 2D Voronoi area by projecting $\mathbf{p}_i$ and its $k$-nearest neighbors to the tangent plane of $\mathbf{p}_i$. 
\citet{jacobson2013winding} showed that the GWN for points that are away from the surface is harmonic, satisfying $\Delta w = 0$. Furthermore, as observed by \citet{xu2023globally}, the GWN produced by a point cloud with randomly oriented normals is nearly zero everywhere (see Figure \ref{fig:pingxing_chuizhi}).

\section{Method}
Our method takes a point cloud $\mathcal{P}=\{\mathbf{p}_i\}_{i=1}^n$, which is sampled from a closed, orientable manifold surface $\partial\Omega$ as input. It begins by assigning a random normal to each point in the cloud. The core of our method lies in the regularization of the generalized winding number $w$ by optimizing an objective function related to the Dirichlet energy of $w$. 
After orientation, we use sPSR~\cite{kazhdan2013screened} to obtain the watertight surface.

\paragraph{Notations}
Throughout the paper, we denote by  $\hat{\mathbf{n}}=\{\hat{\mathbf{n}}_i\}_{i=1}^n$ the set of globally consistent outward unit normal of the surface $\partial \Omega$, and by $\mathbf{n}=\{\mathbf{n}_i\}_{i=1}^n$ an arbitrary unit normal field.
Although the GWN for the input point cloud is computed in a discrete manner, we treat the GWN as a  continuous field for the sake of clarity in our discussion. We define $\Omega^+$ (resp. $\Omega^-$) to be the exterior (resp. interior) of the surface\footnote{To ease discussion, we also define $\Omega^-$, which is the same as the solid $\Omega$, for the interior of the solid object.}. 
The GWN for points $\mathbf{x} \in \partial\Omega$ is defined as $w^{\pm}=\lim_{\epsilon \to 0}w(\mathbf{x}\pm\epsilon \hat{\mathbf{n}})$. We define the consistently oriented normals, $\hat{\mathbf{n}}^+$ and $\hat{\mathbf{n}}^-$, respectively, for the boundary of $\Omega^+$ and $\Omega^-$, where $\hat{\mathbf{n}}^+=-\hat{\mathbf{n}}$ and $\hat{\mathbf{n}}^-=\hat{\mathbf{n}}$. 

\subsection{Boundary Energy}

Since the GWN field associated with a globally consistent orientation is a harmonic function away from the surface and thereby minimizes the Dirichlet energy, \citet{TakayamaJKS14} proposed the minimization of the Dirichlet energy in approximating the GWN from polygonal soups. The Dirichlet energy is defined as an integral over the whole space, excluding the boundary surface $\partial\Omega$, and is given by:
$$
    \min \int_{\mathbb{R}^3 \setminus \partial \Omega} \|\nabla w\|^2 :=\int_{\Omega^+}\|\nabla w\|^2 + \int_{\Omega^-}\|\nabla w\|^2.
$$
A straightforward approach to compute the Dirichlet energy involves subdividing the finite volume around $\partial\Omega$ into sufficiently small elements to accurately capture the variations of the GWN. Subsequently, numerical integration is performed on this discretized space to calculate the Dirichlet energy. In order to calculate integrals accurately, a high-resolution and high-quality tetrahedral mesh is typically necessary, particularly when $\partial\Omega$ has complex geometry and topology. An alternative solution, as proposed by \citet{TakayamaJKS14}, is to convert the volume integral $\int_{\mathbb{R}^3 \setminus \partial \Omega}\|\nabla w\|^2$ into boundary integrals using Green's first identity
\begin{equation}
    \int_{\mathbb{R}^3 \setminus \partial \Omega}\|\nabla w\|^2=\int_{\partial \Omega}(w^- - w^+)\nabla_{\hat{\mathbf{n}}} w^{-},
\label{eqn:boundaryintegralofde}
\end{equation}
where $\nabla_{\hat{\mathbf{n}}}$ denotes the directional derivative along normal direction $\hat{\mathbf{n}}$.

We extend the Dirichlet energy associated with globally consistent normals $\hat{\mathbf{n}}$ to an arbitrary normal vector field $\mathbf{n}$, and define the \textbf{boundary energy} as \begin{equation}f(\mathbf{n}):=\int_{\partial \Omega}(w^- - w^+)\nabla_{\mathbf{n}} w^-.\end{equation}

 Given an arbitrary normal field $\bf n$, we decompose each normal vector $\mathbf{n}$ into two components relative to the globally consistent outward normal $\hat{\mathbf{n}}$ -- a tangential component $\mathbf{n}_{t}$ and a normal component $\mathbf{n}_{n}$, that is, $\mathbf{n} = \mathbf{n}_{n} + \mathbf{n}_{t}$. Leveraging the linearity of directional derivatives, we can express the boundary energy for $\mathbf{n}$ as a sum of its components: $f(\mathbf{n}) = f(\mathbf{n}_{n}) + f(\mathbf{n}_{t})$.

Empirical analysis leads to two key observations: 1) When normals are oriented randomly, both the tangential and normal components of the boundary energy approach zero, yet the Dirichlet energy remains high (Figure \ref{fig:pingxing_chuizhi}, bottom). 2) Conversely, when normals are globally consistent, the tangential component remains negligible $f(\mathbf{n}_{t})\approx 0$, whereas the normal component is large, leading to $f(\mathbf{n}) \approx f(\mathbf{n}_{n})$ (Figure \ref{fig:pingxing_chuizhi}, top). Under this condition, the Dirichlet energy is reduced, aligning with the boundary energy. These observations highlight the competing trends in Dirichlet and boundary energies during the process of transforming random normals to consistent normals. This serves as the inspiration for our approach to directly \textbf{maximize} the boundary energy $f(\mathbf{n})$. 
 As a result, our optimization  becomes 
\begin{equation}
    \max f(\mathbf{n}) := \int_{\partial \Omega}\left(w^- - w^+\right)\nabla_{\mathbf{n}} w^- 
            \label{directE}
\end{equation}

\paragraph{Remark 1}
It is important to note that the boundary energy $f(\mathbf{n})$, despite a superficial similarity to the Dirichlet energy, represents a fundamentally different concept. These two forms of energy coincide only when the normal associated to each point is perpendicular to the surface. For an arbitrary normal field $\bf n$, the two energies diverge. Furthermore, computing the Dirichlet energy for an arbitrary normal field $\bf n$ involves discretizing the space for volume integration - a computationally intensive task. Thus, minimizing the Dirichlet energy with randomly initialized normals is theoretically possible, but it is impractical for real-world applications. Our method overcomes this obstacle by converting volume integration to boundary integration when dealing with random normals, significantly reducing computational demands and eliminating the need for interior discretization.

\subsection{Discretization}
\label{subsec:discretization}

\subsubsection{Sampling GWN Field}
Approximating the boundary integral in Equation (\ref{directE}) using the input points $\{\mathbf{p}_i\}_{i=1}^n$ yields 
$$f(\mathbf{n})\approx \sum\limits_{i=1}^{n}a_i\left(w^-(\mathbf{p}_i)-w^+(\mathbf{p}_i)\right)\nabla_\mathbf{n} w^-(\mathbf{p}_i). 
$$
Given that $w$ and its gradient can be analytically computed using  Equation~\eqref{GWN_calc}, the primary task is to discretize $w^-(\mathbf{p}_i)$ and $w^+(\mathbf{p}_i)$. In our approach, the discretization involves identifying two distinct points $\mathbf{p}^+_i$ and $\mathbf{p}^-_i$ for each input point $\mathbf{p}_i$. These points are strategically positioned such that $\mathbf{p}^+_i$ is located outside $\Omega$ and $\mathbf{p}^-_i$ inside $\Omega$. 
This arrangement ensures that the approximations $w^+(\mathbf{p}_i) \approx w(\mathbf{p}_i^+)$ and $w^-(\mathbf{p}_i) \approx w(\mathbf{p}_i^-)$ hold true.

\begin{figure}
  \centering
  \includegraphics[width=0.475\linewidth]{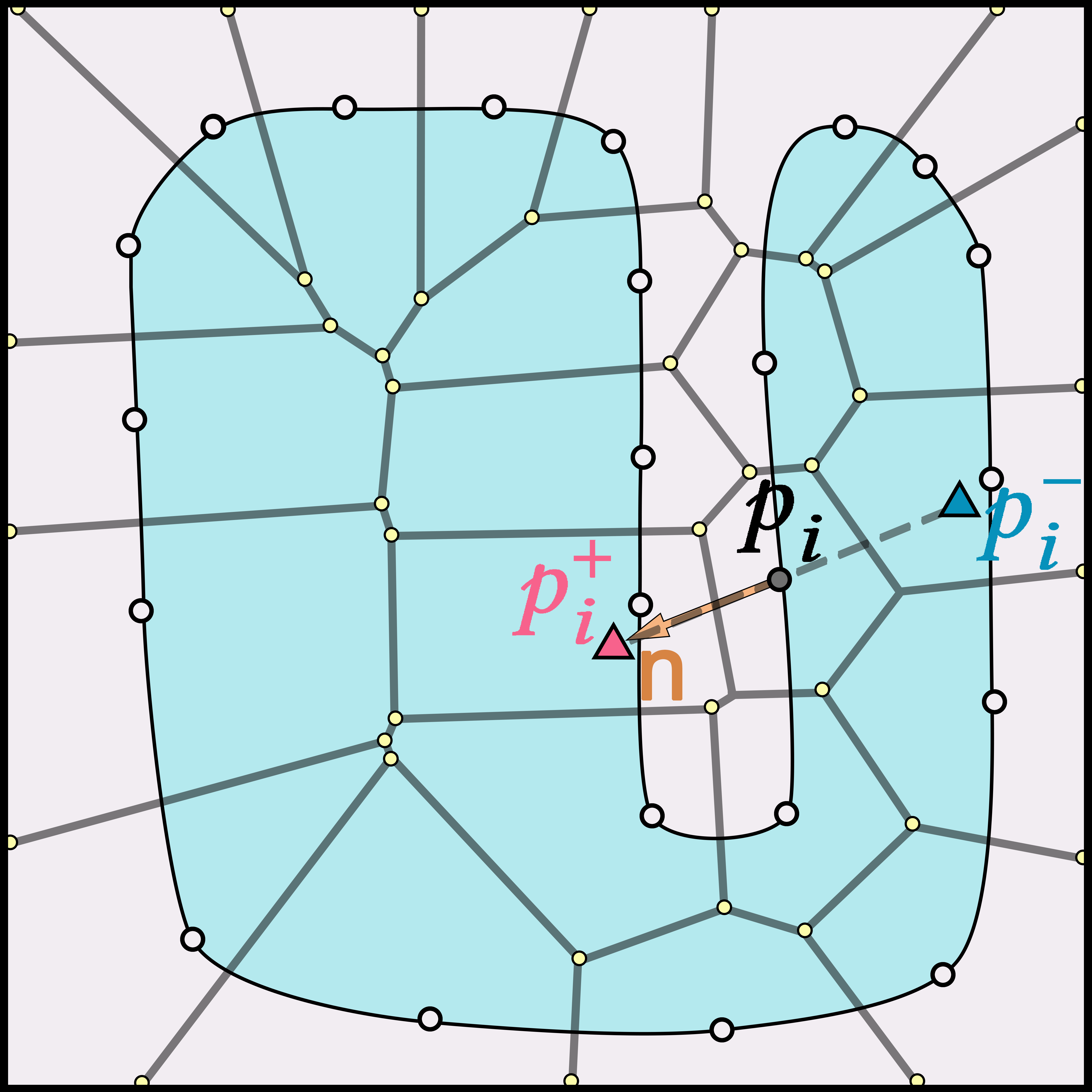}
 \includegraphics[width=0.475\linewidth]{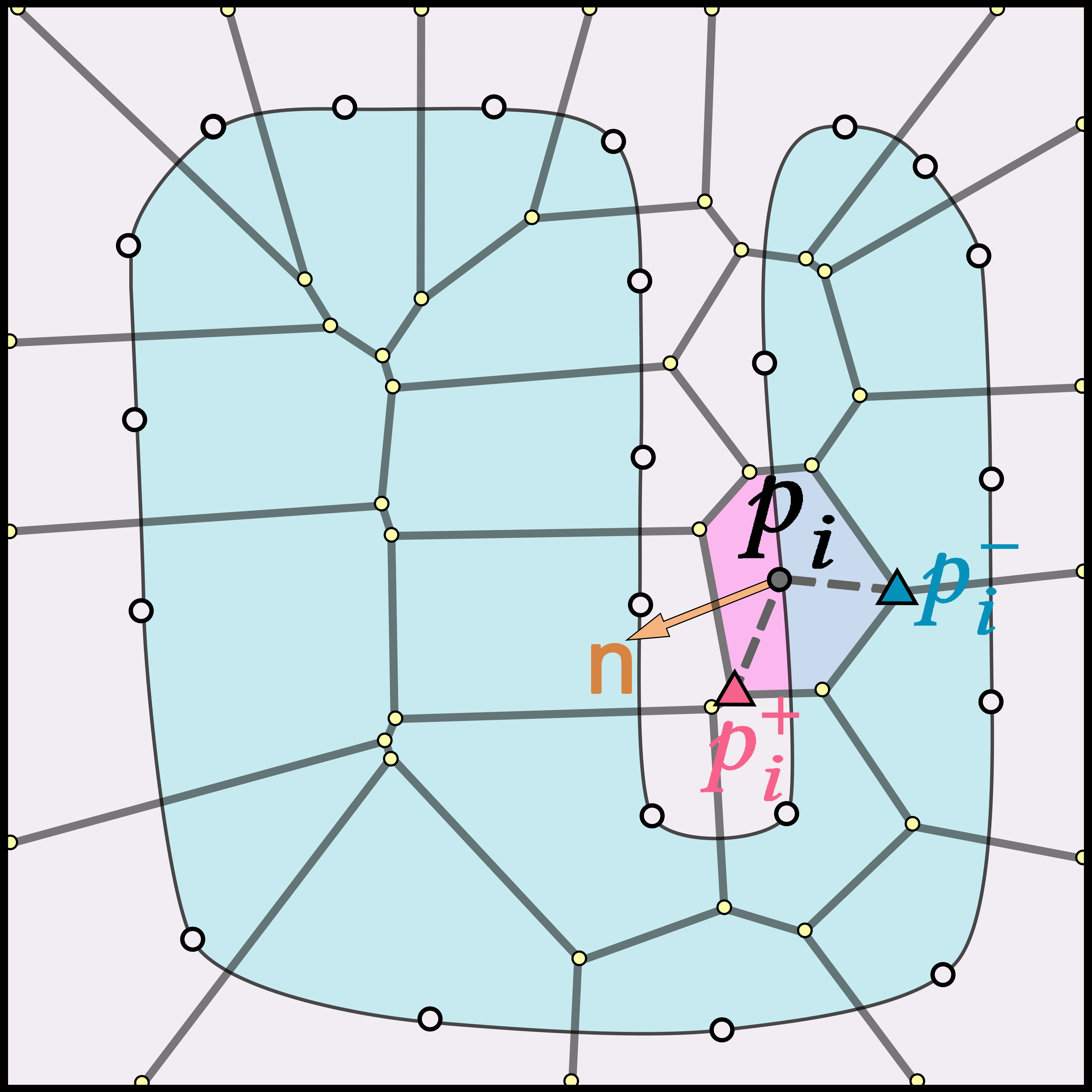}
  \makebox[0.475\linewidth]{(a) Normal displacement}
  \makebox[0.475\linewidth]{(b) Voronoi diagram}\\
  \caption{Illustration of GWN sampling strategies. Input points are represented as large grey circles, with Voronoi vertices shown as small yellow circles. 
  The interior $\Omega^-$ is shaded blue and the exterior is in gray. A representative point $\mathbf{p}_i$ is highlighted in black, with its associated samples $\mathbf{p}_i^+$ and $\mathbf{p}_i^-$ shown as red and blue triangles, respectively. (a) The normal displacement strategy, though straightforward and easy to implement, is ineffective for models with thin structures. In this example, it positions both $\mathbf{p}_i^+~(=\mathbf{p}_i+\mathbf{n}_i)$ and $\mathbf{p}_i^-~(=\mathbf{p}_i-\mathbf{n}_i)$ on the \textbf{same} side of the target surface (specifically, both inside the surface). (b) Using Voronoi diagrams, we identify $\mathbf{p}_i^\pm$ by selecting two Voronoi vertices associated with $\mathbf{p}_i$ that best align with the current normal $\mathbf{n}_i$. This approach ensures that for the \textbf{majority} of the input points, their chosen $\mathbf{p}^\pm$ are consistently positioned on \textbf{opposite} sides of $\partial\Omega$, thereby providing a reliable sampling of the GWN field around the target surface.} 
  \label{fig:sampling}
\end{figure}

A straightforward approach for identifying $\mathbf{p}_i^{\pm}$ is to set $\mathbf{p}_i^{\pm}=\mathbf{p}_i \pm \epsilon\mathbf{n}_i^{'}$, where $\mathbf{n}_i^{'}$ denotes the current {estimate of the} normal and $\epsilon~(> 0)$ is a  predetermined value. However, this strategy does not reliably ensure that $\mathbf{p}_i^{-}$ and $\mathbf{p}_i^+$ are located inside and outside the surface, respectively, especially in cases involving complex geometries and thin structures (see Figure~\ref{fig:sampling} (a)). 

To avoid these issues, we opt for Voronoi vertices to discretize $\mathbf{p}^{\pm}$. Given that many Voronoi cells are infinite, we truncate them using an enlarged bounding box of the input point cloud $\mathcal{P}$. Both the Voronoi vertices and the vertices along the clipped boundary are considered as potential candidates for $\mathbf{p}^{\pm}$. This strategy is supported by two key observations: Firstly, most Voronoi vertices, although not necessarily far away from the point cloud\footnote{As pointed out in \citet{Voronoi_based_reconstruction}, in two dimensions, the Voronoi vertices of a dense set of sample points on a curve approximate the medial axis of the curve. However, this is not necessarily true in three dimensions.}, tend to be on opposite sides of the target surface in practice, thereby providing suitable candidate positions for $\mathbf{p}^\pm$. 
Secondly, as demonstrated in \cite{xu2023globally}, Voronoi vertices exhibit considerable robustness against noise so that we can use them to distinguish interior and exterior. Computational results further confirm that for models with low levels noise, employing Voronoi vertices yields a reliable sampling of the GWN field around the surface. For a more detailed discussion, refer to the supplementary material. 
In our implementation, for each $\mathbf{p}_i$, we identify $\mathbf{p}_i^+$ and $\mathbf{p}_i^-$ as the two Voronoi vertices of the Voronoi cell associated with $\mathbf{p}_i$ that best align with the current normal $\mathbf{n}_i$ (see Figure~\ref{fig:sampling} (b)). 

\paragraph{Remark 2} At the beginning of the optimization process with randomly initialized normals, some points $\mathbf{p}_i$ may experience inside-out flipping, where $\mathbf{p}_i^+$ is positioned inside and $\mathbf{p}_i^-$ outside. This inside-out configuration does not hinder our boundary energy maximization, as the samples $\mathbf{p}_i^+$ and $\mathbf{p}_i^-$ still remain on \textbf{opposite sides} of the target surface. Computational results show that the energy maximization process effectively reduces the frequency of such inside-out cases. For example, on the Bunny model, after 10 iterations, all $\mathbf{p}_i^+$ are consistently positioned outside and all $\mathbf{p}_i^-$ inside. See Figure 4 in the supplementary material.

\subsubsection{Energy Maximization}
Using the Voronoi vertices to sample the GWN field around $\partial\Omega$, we can express the maximization problem in Equation~\eqref{directE} as: 
\begin{displaymath}
    \max  \sum\limits_{i=1}^{n}a_i\left(w(\mathbf{p}_i^{-}) - w(\mathbf{p}_i^{+})\right)\nabla_{\mathbf{n}_i} w(\mathbf{p}_i)
\end{displaymath}

\begin{figure}
  \centering
\includegraphics[width=1.0\linewidth]{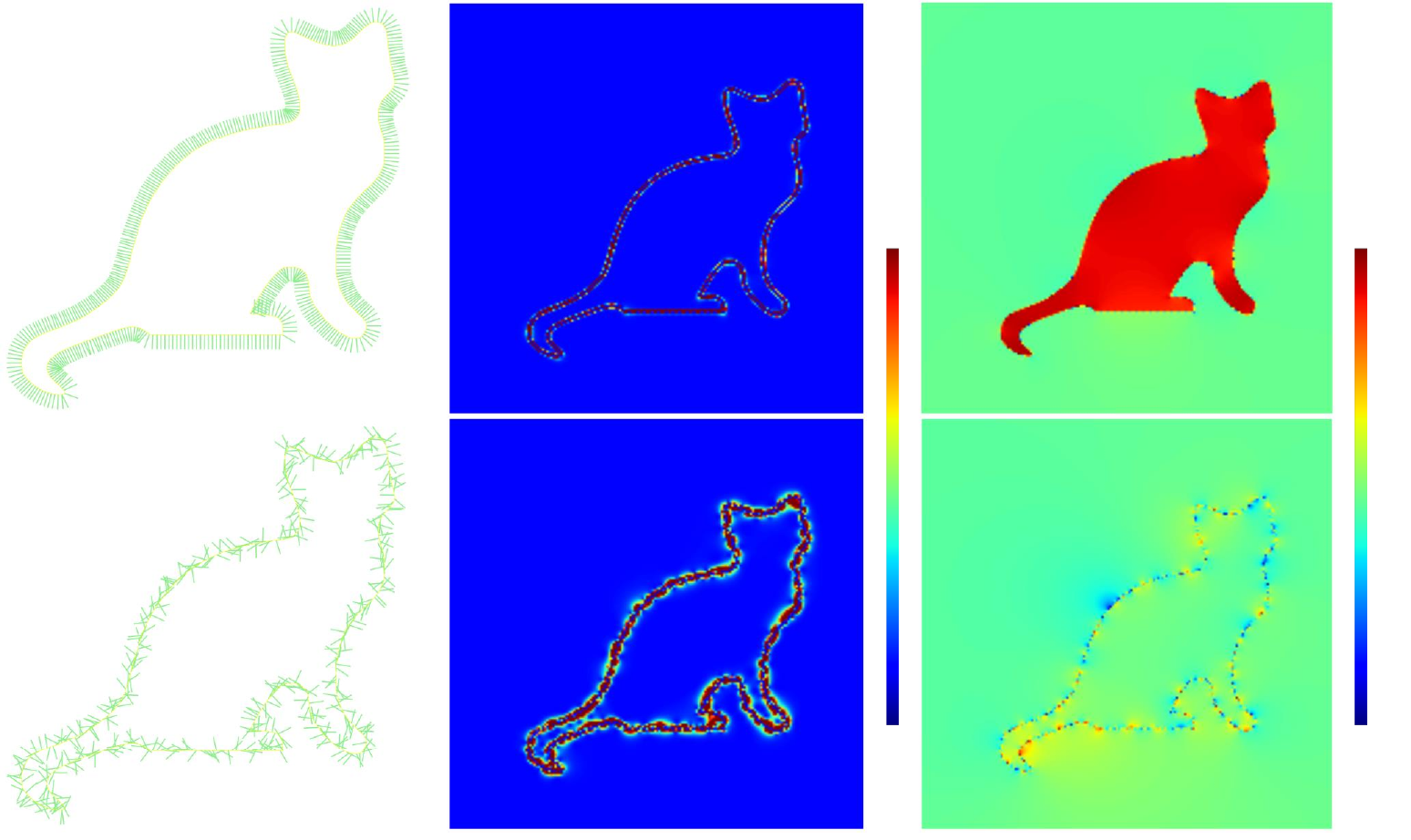}
\put(-91, 19){\makebox(0,0)[tl]{\small \textsf{<5}}}
\put(-92, 110.5){\makebox(0,0)[tl]{\small \textsf{>}$10^2$}}
\put(-5, 21){\makebox(0,0)[tl]{\small \textsf{-1}}}
\put(-5, 60){\makebox(0,0)[tl]{\small \textsf{0}}}
\put(-5, 102){\makebox(0,0)[tl]{\small \textsf{1}}}\\
\makebox[1.0in]{$\mathbf{n}$}
\makebox[1.0in]{$\|\nabla w\|^2$}
\makebox[1.0in]{$w$}
\caption{Visualization of the GWN field $w$ and its gradient $\|\nabla w\|^2$ for globally consistent normals (top) and random normals (bottom). The decomposed boundary energies $f(\mathbf{n}_{n})$ and $f(\mathbf{n}_{t})$ are (167.53, -3.21$\times 10^{-15}$) and ($4.77\times 10^{-6}$, $2.11\times 10^{-5}$), respectively.}
 \label{fig:pingxing_chuizhi}
\end{figure}
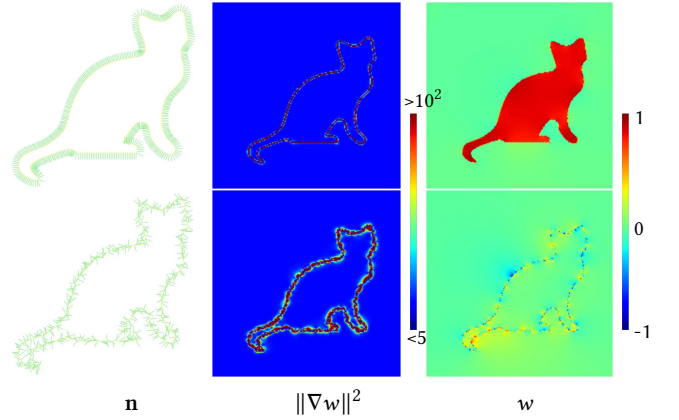

\begin{figure}[!htbp]
  \centering
  \hspace{0.3in}
  \includegraphics[width=0.9\linewidth]{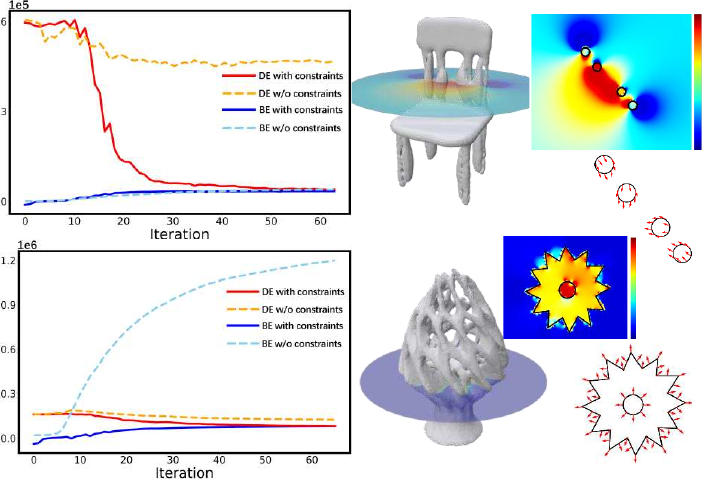}
      \put(0,138.5){\makebox(0,0)[bl]{\small \textsf{1}}}
    \put(0.5,119){\makebox(0,0)[bl]{\small \textsf{0}}}
    \put(-2.5,100){\makebox(0,0)[bl]{\small \textsf{-1}}}
    \put(-77,77){\makebox(0,0)[bl]{\small \textsf{Case 1}}}
    \put(-21,68){\makebox(0,0)[bl]{\small \textsf{2}}}
    \put(-21,56){\makebox(0,0)[bl]{\small \textsf{1}}}
    \put(-20.6,44){\makebox(0,0)[bl]{\small \textsf{0}}}
    \put(-77,2){\makebox(0,0)[bl]{\small \textsf{Case 2}}}
  \caption{The ambiguities of jump boundary conditions in Case 1 and Case 2. For each case, we plot the boundary energy (BE) and the Dirichlet energy (DE) with and without the constraints $0\leq w\leq 1$. By restricting the range of the winding numbers, DE and BE converge when the associated normals are globally consistent. We also visualize the GWN field using cut views and indicate the normals for the cross sections. } 
  \label{fig:twocase}
\end{figure}
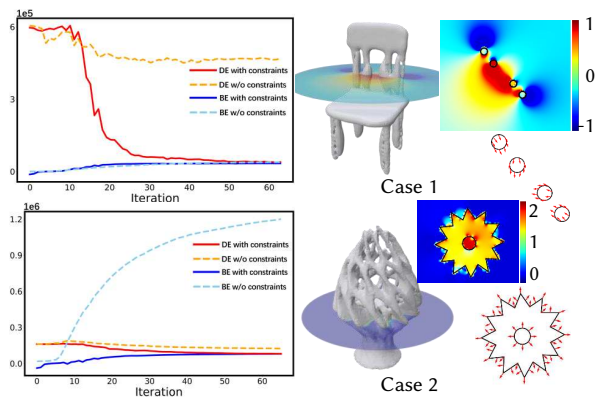

Directly maximizing the above boundary energy often results in normals that, while perpendicular to the boundary surface, are not globally consistent. The inconsistency is mainly due to ambiguities associated with the jump boundary conditions. 
Figure \ref{fig:twocase} shows two common types of ambiguities encountered in 3D space: Case 1, where $w^+ = -1$ and $w^- = 0$, and Case 2, where $w^+ = 1$ and $w^- = 2$.
A direct approach to resolving the ambiguity in jump boundary conditions involves constraining the winding numbers to fall within the range of 0 and 1, that is:
\begin{displaymath}
 0 \le w(\mathbf{p}_i^{\pm}) \le 1, \quad i = 1, 2, \ldots, n. 
\end{displaymath}
\begin{algorithm} 
    \small
    \fontsize{9}{8}\selectfont
    \caption{\small Globally consistent point orientation via optimizing the boundary energy of GWN} 
    \begin{algorithmic}
		\STATE Input: Point cloud $\mathcal{P} = \{\mathbf{p}_i\}_{i=1}^n$
  \STATE Output: Globally consistent outward unit normal for each point

     \STATE Compute the 3D Voronoi diagram with $\{\mathbf{p}_i\}$ as seeds
     
     \FOR{each point $\mathbf{p}_i\in\mathcal{P}$}
                 \STATE Assign a random normal $\mathbf{n}_i=(\cos u_i \sin v_i, \sin u_i\sin v_i, \cos v_i)^{T}$
            
             \STATE Calculate the geodesic Voronoi area $a_i$ 
          
            \STATE Collect the Voronoi vertices $\{\mathbf{q}_i^k\}_{k=1}^{m_i}$ 
    \ENDFOR
    \WHILE{the L-BFGS solver does not converge}
        \FOR{each point $\mathbf{p}_i\in\mathcal{P}$}
            \STATE Identify $\mathbf{p}_i^+ = \arg\min\limits_{\{\mathbf{q}_{i}^{k}\}}  \angle(\mathbf{q}_{i}^{k}-\mathbf{p}_i, \mathbf{n}_i^{'})$ 
            \STATE Identify $\mathbf{p}_i^- = \arg\max\limits_{\{\mathbf{q}_{i}^{k}\}} \angle(\mathbf{q}_{i}^{k}-\mathbf{p}_i, \mathbf{n}_i^{'})$ 
            \STATE Calculate $w(\mathbf{p}_i^+)$ and $w(\mathbf{p}_i^-)$ as in Equation~\eqref{GWN_calc}
            \STATE Calculate $\nabla w(\mathbf{p}_i) = \sum\limits_{j=1,j\neq i}^{n} a_i\nabla P(\mathbf{p}_i, \mathbf{p}_j)$
           
        \ENDFOR
         \STATE Calculate the boundary energy as in Equation~\eqref{FinalE} and its gradient
        \STATE Update $\{u_i, v_i\}_{i=1}^{n}$ using the L-BFGS solver
    \ENDWHILE 
	\end{algorithmic} 
    \label{algo_pipeline}
\end{algorithm}

However, this method introduces $2n$ inequality constraints, significantly increasing the complexity of the optimization problem. Instead, we propose not to directly restrict the value of $w$ using the above inequality constraints. Instead, we 1) replace the GWN difference $w^{-}-w^{+}$ in Equation~\eqref{directE}
with $|w^{-}|-|w^{+}|$, which encourages $w\geq 0$, thereby penalizing Case 1; and 2) add an additional term $$g\left(w(\mathbf{p}_i^{\pm})\right)=\left(1-e^{\max\{w(\mathbf{p}_i^{+}), 1\}}\right) +\left(1- e^{\max\{w(\mathbf{p}_i^{-}), 1\}}\right),$$which encourages $w\leq 1$, thereby penalizing Case 2. 
 With these changes, the final energy becomes an \textbf{unconstrained} optimization problem:
\begin{equation}
    \max \sum\limits_{i=1}^{n}a_i\left(|w(\mathbf{p}_i^{-})| - |w(\mathbf{p}_i^{+})|\right)\nabla_{\mathbf{n}_i} w(\mathbf{p}_i) + g\left(w(\mathbf{p}_i^{\pm})\right).
    \label{FinalE}
\end{equation}

We solve this optimization using the L-BFGS solver\footnote{https://github.com/yixuan/LBFGSpp}. During each iteration, we examine the current normal $\mathbf{n}_i$. Through a linear scan, we determine $\mathbf{p}_i^+$ and $\mathbf{p}_i^-$, which correspond to the Voronoi vertices forming the smallest and largest angles, respectively, with $\mathbf{n}_i$.
To force the normals to be unit length, we parameterize each normal as $\mathbf{n}_i=(\cos u_i \sin v_i, \sin u_i \sin v_i, \cos v_i)^T$, where $u_i$ and $v_i$ serve as the variables in the objective function Equation~\eqref{FinalE}.

\paragraph{Remark 3} Our Voronoi diagram strategy does not guarantee that the selected $\mathbf{p}^\pm$ will always be on opposite sides of the target surface, experimental results suggest that most input points do not face the issue of $\mathbf{p}^\pm$ being on the same side. In the rare occasions when  $\mathbf{p}_i^+$ and $\mathbf{p}_i^-$ do end up on the same side, our optimization method effectively addresses these instances by penalizing them. Specifically, when $\mathbf{p}_i^+$ and $\mathbf{p}_i^-$ are on the same side, the value of $|w(\mathbf{p}_i^-)| - |w(\mathbf{p}_i^+)|$ reduces, leading to lower energy. Our objective is to maximize boundary energy, and with the  normals updated, it is unlikely that these same-side samplings will recur in subsequent iterations.
\paragraph{Remark 4} \label{time_complex} In our current implementation, we calculate $\nabla w$ and $w$ in a brute-force manner. Let $n$ be the number of input points in $\mathcal{P}$ and $m$ represent the number of Voronoi vertices. Each iteration begins with an $O(n+m)$ traversal of Voronoi vertices to identify \textbf{all} $\mathbf{p}^\pm$s. Subsequently, for each input point, computing $\nabla w$ and $w$ via the Poisson kernel requires $O(n)$ time. Therefore, the computational cost per iteration is $O(n^2+n+m)$. 
To enhance runtime efficiency, one possible strategy is to accelerate the computation using the fast multipole method~\cite{GREENGARD1987325}. This enables querying neighboring points in $O(\log n)$ time, potentially reducing the overall time complexity per iteration to $O(n\log n)$. 

\begin{figure}
  \centering
  \includegraphics[width=1.0\linewidth]{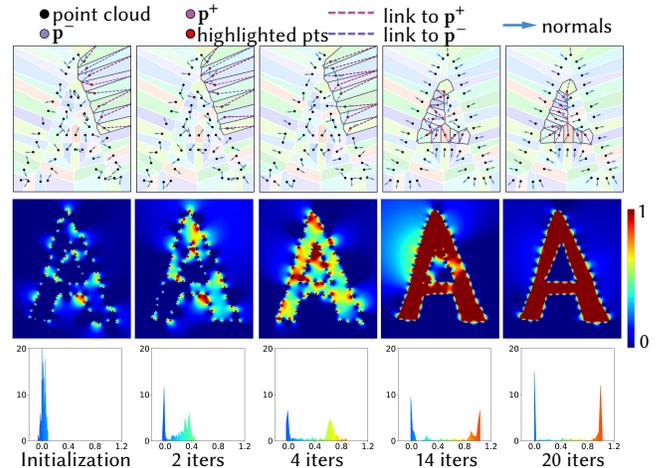}
      \put(-225,162){\makebox(0,0)[bl]{\small \textsf{point cloud}}}
    \put(-225,155){\makebox(0,0)[bl]{\small \textsf{$\mathbf{p}^-$}}}
    \put(-170,162){\makebox(0,0)[bl]{\small \textsf{$\mathbf{p}^+$}}}
    \put(-170.5,154){\makebox(0,0)[bl]{\small \textsf{highlighted pts}}}
    \put(-100,161){\makebox(0,0)[bl]{\small \textsf{link to $\mathbf{p}^+$}}}
    \put(-100,154){\makebox(0,0)[bl]{\small \textsf{link to $\mathbf{p}^-$}}}
    \put(-40,159){\makebox(0,0)[bl]{\small \textsf{normals}}}
    \put(-237,-5){\makebox(0,0)[bl]{\small \textsf{Initialization}}}
    \put(-180,-5){\makebox(0,0)[bl]{\small \textsf{2 iters}}}
    \put(-135,-5){\makebox(0,0)[bl]{\small \textsf{4 iters}}}
    \put(-88,-5){\makebox(0,0)[bl]{\small \textsf{14 iters}}}
    \put(-40,-5){\makebox(0,0)[bl]{\small \textsf{20 iters}}}
    \put(-3,40){\makebox(0,0)[bl]{\small \textsf{0}}}
    \put(-4,89){\makebox(0,0)[bl]{\small \textsf{1}}}
  \caption{Illustration of our algorithm on a 2D toy model. In different iterations, we visualize the normals, $\mathbf{p}^+$, $\mathbf{p}^-$ (row 1), winding number field (row 2), $w(\mathbf{p}^+)$ and $w(\mathbf{p}^-)$ (row 3). We highlight the regions with significant variations in normals between adjacent iterations using gray lines and red points. 
  }
  \label{fig:A_all}
\end{figure}

\subsection{Contrasting Our Method with Previous Approaches}
\label{subsec:comparison}
Our method, while sharing similarities with the works of \citet{TakayamaJKS14}, PGR~\cite{lin2022surface}, and GCNO~\cite{xu2023globally}, particularly in their use of the generalized winding number field, diverges significantly in several key aspects.

\paragraph{Ours vs. \citet{TakayamaJKS14}.'s method}
Although both methods aim at optimizing objective functions that are related to to the Dirichlet energy, there are two key differences that set our approach apart:
1) Takayama et al. primarily addressed scenarios where point normals are perpendicular to the boundary surface yet inconsistent. In such cases, we observe a significant increase in boundary energy (Figure~\ref{fig:pingxing_chuizhi}). However, when normals are initialized randomly, the boundary energy tends to diminish to nearly zero, indicating a fundamental difference in the objective functions.
2) Their approach relies on a brute-force solver, exhaustively searching for the optimal normal configuration. As a result, its applicability is confined to small and simple models. In contrast, our method utilizes an L-BFGS solver.

\paragraph{Ours vs. GCNO~\cite{xu2023globally}}\label{vs_gcno} Our approach fundamentally differs from GCNO in terms of methodology, discretization techniques and time complexity. 1) \textbf{Methodology}: Unlike GCNO, which regularizes the GWN of Voronoi vertices to exact values of 0 and 1, our method focuses on the boundary energy by evaluating the GWN difference $|w(\mathbf{p}_i^-)|-|w(\mathbf{p}_i^+)|$ and the GWN gradient. This strategy does not depend on precise GWN values, offering a more flexible approach. 2) \textbf{Discretization}:  GCNO assumes a uniform distribution of Voronoi vertices both inside and outside the object—an assumption that is less reliable for models with complex geometries. Figure~\ref{fig:GCNO_all_fail} shows an example where GCNO is unable to optimize its objective function when the assumption does not hold. Our method, by contrast, does not rely on such an assumption, making it adaptable to a broader range of 3D models. 3) \textbf{Time complexity}: GCNO requires computing $w$ for \textbf{all} Voronoi vertices associated with each point in the input point cloud. Our method, however, requires computing  $\nabla w$ for $\mathbf{p}$ and $w$ for two specific points, $\mathbf{p}^\pm$, within the Voronoi cell. 
This leads to significantly lower computational demands in comparison to GCNO. Specifically, GCNO has a time complexity $O(nm)$ for each iteration, while ours is $O(n^2+n+m)$ per iteration. Given that the number of input points $n$ is typically much lower than the number of Voronoi vertices $m$ ($n\ll m$), as exemplified by the Bunny model ($n = 8,830$, $m = 224,085$), our method outperforms GCNO in terms of scalability and runtime performance. Experimental results indicate that GCNO is suited for models with up to 10K points, typically generating within an hour. However, for larger models containing 40K points, GCNO fails to complete within 24 hours. In contrast, our method can deliver results in just three hours. See the supplementary material for the details of the scalability test.

\paragraph{Ours vs. PGR~\cite{lin2022surface}} PGR focuses on regularizing the GWN directly at the input points. In the context of noisy point clouds, achieving a GWN value of exactly 0.5 for these points is challenging. Unlike PGR, our method does not constrain GWN values. Instead we regularize the differences and gradients of the GWN both inside and outside of the input model, significantly enhancing our method's robustness to noise. Furthermore, PGR requires solving a dense linear system, restricting its applicability to small models. In contrast, our method employs an L-BFGS solver to optimize the boundary energy, offering a solution that is both fast and memory-efficient.

\begin{table*}
  \centering
  \resizebox{\linewidth}{!}{
  \fontsize{4}{3}\selectfont 
  \begin{tabular}{c|cccccc}
  \toprule
    \specialrule{-1pt}{1pt}{1pt}
    Models (\# points, Fig.)
    & $\text{Dipole}^{*}$ & $\text{PGR}^{*}$ & iPSR & $\text{NeuralGF}^{*}$ & GCNO & Ours \\
    \specialrule{-1pt}{1pt}{1pt}
    \midrule
    \specialrule{-1pt}{1pt}{1pt}
    
    BS (4000, supp.)
    & 4.5/5.7/0.17/28/2.0 & 17/21/0.12/\textbf{2.0}/0.50 & 3.7/4.5/\textbf{0.11}/9.7/0.094 & 5.1/6.0/0.15/640/1.9 & 8.1/9.9/0.21/810/\textbf{0.067} & \textbf{3.6/4.3/0.11}/95/0.075 \\
    
    Bird (8187, Fig.~\ref{fig:NGLO_SHSNET}) 
    & 12/37/1.7/50/1.9 & 7.4/11/0.063/\textbf{8.1}/0.80 & 2.2/5.1/0.065/14/0.10 & 5.2/8.8/0.069/640/1.9 & 89/98/28/2800/0.14 & \textbf{2.0/3.4/0.060}/360/\textbf{0.084} \\

    linkCupTop (8602, Fig.~\ref{fig:NGLO_SHSNET})
    & 89/120/3.6/60/2.2 & 23/28/0.27/\textbf{6.7}/0.85 & 24/28/0.20/25/0.11 & 87/110/3.4/640/1.9 & 89/110/3.3/2200/0.14 & \textbf{12/15/0.19}/400/\textbf{0.091} \\

    BunnyPeel (8648, Fig.~\ref{fig:sparse_compare})
    & 51/82/25/53/2.0 & 22/26/0.099/\textbf{10}/0.85 & 12/17/0.099/18/0.10 & 48/70/11/640/1.9 & 45/63/36/2400/0.14 & \textbf{7.2/12/0.094}/400/\textbf{0.092} \\

    Horse (8657, supp.)
    & 8.2/23/0.11/51/2.4 & 13/17/\textbf{0.065}/\textbf{7.1}/0.85 & 3.1/5.0/0.070/14/0.10 & 6.6/9.5/\textbf{0.065}/640/1.9 & 8.2/9.9/0.072/2500/0.14 & \textbf{2.9/4.2/0.065}/400/\textbf{0.085} \\

    397horse (8784, supp.) 
    & 8.9/28/0.19/50/2.4 & 15/21/0.066/\textbf{7.0}/0.86 & 3.1/5.0/0.063/15/0.10 & 8.1/17/0.45/640/1.9 & 8.2/9.9/0.064/2700/0.14 & \textbf{2.9/4.2/0.062}/400/\textbf{0.086} \\

    Bunny (8830, supp.)
    & 6.7/9.2/0.13/62/2.1 & 15/20/0.14/\textbf{7.3}/0.86 & 4.9/6.9/\textbf{0.11}/13/0.11 & 5.4/6.6/0.17/640/1.9 & 8.9/11/0.16/2700/0.14 & \textbf{3.1}/\textbf{4.1}/0.12/410/\textbf{0.094} \\

    Cup (10130, supp.)
    & 4.4/19/0.14/60/2.0 & 11/14/\textbf{0.13}/\textbf{9.7}/1.0 & \textbf{1.5}/2.5/0.15/11/0.11 & 3.9/4.5/\textbf{0.13}/640/1.9 & 7.4/9.5/\textbf{0.13}/3700/0.16 & 1.7/\textbf{2.2/0.13}/550/\textbf{0.088} \\

    Botijo (11305, supp.)
    & 14/37/0.23/63/2.4 & 16/20/0.080/\textbf{13}/1.1 & 3.0/\textbf{4.9}/0.077/15/0.12 & 15/38/2.3/640/1.9 & 35/64/54/4100/0.19 & \textbf{2.6/3.7/0.077}/660/\textbf{0.086} \\
    
    30cup (11311, supp.)
    & 4.4/15/0.090/65/2.4 & 14/16/0.091/14/1.1 & \textbf{1.7}/3.1/0.093/\textbf{13}/0.12 & 32/64/97/640/1.9 & 9.2/11/0.097/4400/0.17 & 2.1/\textbf{3.0/0.086}/670/\textbf{0.095} \\

    Candle (12221, Fig.~\ref{fig:sparse_compare})
    & 65/93/1.12/67/2.4 & 42/53/0.97/\textbf{18}/1.2 & 18/25/\textbf{0.099}/34/0.14 & 30/48/11/640/1.9 & 50/72/36/7900/0.20 & \textbf{14}/26/\textbf{0.094}/790/\textbf{0.089} \\

    Elk (12541, Fig.~\ref{fig:sparse_compare})  
    & 17/48/7.8/77/2.1 & 19/30/0.17/22/1.3 & 3.0/6.8/0.12/\textbf{18}/0.12 & 17/43/5.9/640/1.9 & 48/83/20/7400/0.20 & \textbf{2.9/4.6/0.11}/1000/\textbf{0.10} \\

    Lion (13138, Fig.~\ref{fig:sparse_compare})
    & 14/28/0.57/71/3.2 & 21/28/0.061/16/1.4 & 9.1/14/0.068/\textbf{14}/0.11 & 12/17/0.068/640/1.9 & 19/31/0.20/5100/0.21 & \textbf{6.5/11/0.057}/900/\textbf{0.086} \\

    Vase (13375, Fig.~\ref{fig:NGLO_SHSNET})
    & 50/84/8.5/81/2.5 & 26/33/0.26/\textbf{18}/1.4 & 12/16/\textbf{0.23}/39/0.13 & 77/104/5.3/640/1.9 & 88/114/11/6200/0.21 & \textbf{5.8/8.1/0.23}/920/\textbf{0.099} \\

    Fertility (13802, supp.)
    & 52/94/120/76/3.2 & 15/20/\textbf{0.073}/20/1.5 & 2.3/4.0/0.10/\textbf{19}/0.14 & 5.0/6.3/0.081/640/1.9 & 8.0/11/0.091/10000/0.22 & \textbf{2.2}/\textbf{3.2}/0.075/970/\textbf{0.088} \\
    
    Fandisk (13854, supp.)
    & 4.7/9.9/0.13/79/2.2 & 18/25/0.12/22/1.5 & \textbf{2.6}/5.3/0.12/\textbf{17}/0.16 & 4.6/6.0/0.18/640/1.9 & 9.5/12/\textbf{0.11}/5500/0.20 & 2.8/\textbf{4.3/0.11}/990/\textbf{0.092} \\
    
    Felinei (15057, supp.)
    & 36/72/25/81/3.0 & 19/24/0.092/23/1.6 & 5.7/10/0.088/\textbf{18}/0.11 & 30/57/16/640/1.9 & 89/97/17/8500/0.24 & \textbf{4.1/7.2/0.084}/1200/\textbf{0.092} \\

    Pulley (16116, supp.)
    & 13/31/0.12/86/3.1 & 26/33/0.092/23/1.8 & 5.0/\textbf{7.5}/0.14/16/0.16 & 12/17/0.090/640/1.9 & 7.4/8.8/0.12/12000/0.22 & \textbf{3.8/5.3/0.083}/1300/\textbf{0.092} \\

    \hline
    Average
    & 25/46/11/64/2.4 & 19/25/0.16/\textbf{14}/1.1 & 6.5/9.5/\textbf{0.11}/18/0.12 & 22/35/8.0/640/1.9 & 35/46/9.5/5100/0.17 & \textbf{4.6/7.0/0.11}/690/\textbf{0.090} \\
    
    \specialrule{-1pt}{1pt}{1pt}
    \bottomrule
  \end{tabular}
  }
  \caption{Quantitative results on clean point clouds. We present five metrics: mean and standard deviation of angular discrepancies (degree), CD ($10^{-3}$), runtime (seconds), and GPU or CPU memory utilization (MB). * indicates GPU-based. The best results are emphasized in bold. See results on noisy point clouds in the supplementary material. 
  }
  \label{table:noise_free}
\end{table*}

\begin{figure*}[!htbp]
  \centering
\includegraphics[width=1.0\linewidth]{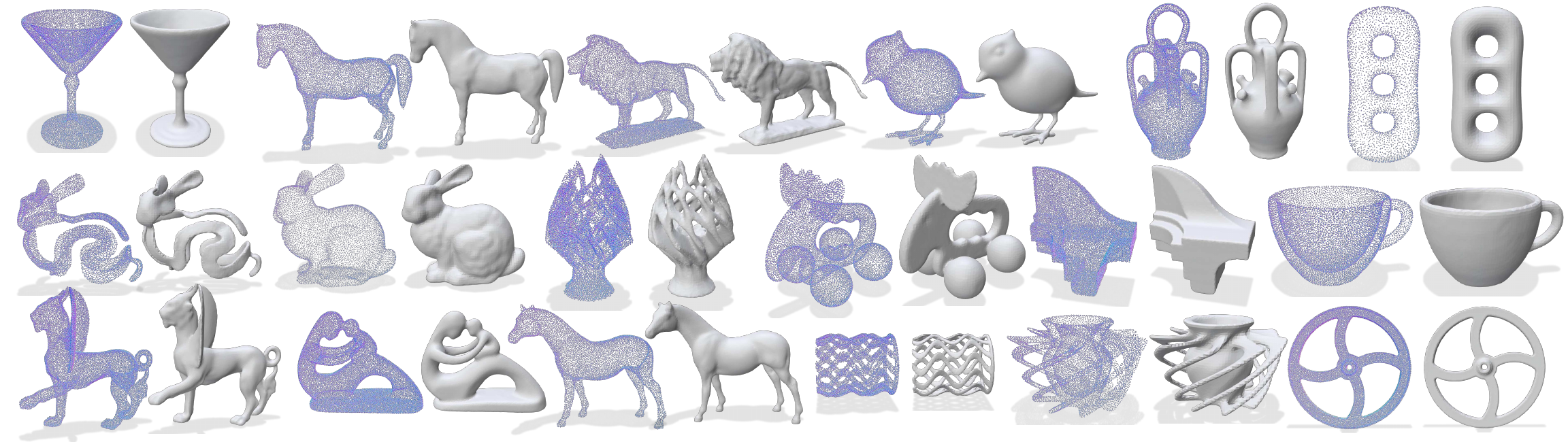}
  \caption{Reconstruction results for the 18 test models. }
  \label{fig:allmodels}
\end{figure*}
\section{Experimental Results}
\paragraph{Experimental setup} We perform our experiments on a PC equipped with an i7-12700H CPU and 16GB of RAM. The GPU-based approaches \cite{lin2022surface, metzer2021orienting, li2023neuralgf} were executed on a single RTX 3090. We evaluate our method using 18 models, with diverse geometry and topology (see Figure~\ref{fig:allmodels}). All point clouds are uniformly scaled to fit within the cube $[-1, 1]^3$. Noise was synthetically generated using a standard Gaussian distribution with $\mu=0.0$ and $\sigma^2=1.0$, introducing random displacements to each point. We modulate the noise intensity to 0.5\% and 0.75\% of the diagonal length of the bounding box. Our method involves a single parameter: the neighborhood size, which is used for approximating the Voronoi area, $a_i$, in the discretization of the GWN. Despite the diversity in model geometries, we use a neighborhood size of 15 across all tests. We use the L-BFGS solver for optimizing the unconstrained boundary energy. After establishing the point orientations, we use screened Poisson surface reconstruction \cite{kazhdan2013screened} on the oriented point clouds, generating watertight surfaces. We compare our method with five state-of-the-art approaches GCNO~\cite{xu2023globally}, Dipole~\cite{metzer2021orienting}, PGR~\cite{lin2022surface}, NeuralGF~\cite{li2023neuralgf}, and iPSR~\cite{hou2022iterative}, utilizing the official implementations provided by the authors along with their default parameter settings.
We assess the quality of orientation and reconstruction by analyzing the angular discrepancy between the predicted normals and the ground-truth normals, along with the Chamfer Distance (CD), which measures the closeness between the ground-truth surface and the reconstructed surface. To quantitatively describe this distribution, we report the mean $\mu$ and the standard deviation $\sigma$ of the angle differences in degree. The smaller the metrics, the higher the consistency of the predicted normals. We also document the running time and the peak memory consumption for computational efficiency.

\paragraph{Orientation \& reconstruction results} We present the performance of various methods across 18 noise-free models in Table \ref{table:noise_free}, and provide results under noisy conditions in the supplementary material. In the noise-free scenarios, our method achieves the best scores in $\mu$, $\sigma$ and CD for 83\%, 88\%, and 88\% of the models, respectively.
Our method demonstrates resilience to low level noise, as it is a global method that can capture the normal consistency from a global perspective.
See Figures~\ref{fig:sparse_compare} and~\ref{fig:teaser}   for the comparison of the reconstructions. Our method is particularly effective in processing models characterized by complex topology and thin structures, areas where competing methods frequently encounter difficulties. 

\paragraph{Runtime}
As analyzed in Section~\ref{subsec:discretization}, our method's time complexity is $O(n^2 + n + m)$ per iteration. Typically, $m$ increases linearly with $n$ for common models. However, it is important to note that in the worst-case scenario, $m$ can scale quadratically with $n$.  Consequently, our method is faster than GCNO, which operates with a time complexity of $O(nm)$ per iteration. Computational results demonstrate that our method is 5-10 times faster than GCNO across 18 test models, with an average speed-up factor of 7.3. 

\paragraph{Space complexity} Our algorithm has space complexity of $O(n+m)$, since it only needs to store the input point clouds and its associated Voronoi diagram. Both PGR~\cite{lin2022surface} and VIPSS~\cite{VIPSS2019} necessitate solving dense linear systems, resulting in a space complexity of $O(n^2)$. This quadratic space complexity constrains them to smaller-scale models, generally limited to those with up to 10K points. In contrast, our method scales up to 100K points.

\paragraph{Comparison with supervised methods} We also compare our method with recent supervised approaches, NGLO~\cite{li2023neuralgf} and SHS-Net~\cite{li2023shsnet}. NGLO operates in two stages: The initial phase, which is unsupervised, produces a set of normals. While globally consistent, these  may not always be perpendicular to the surface. This is followed by a supervised module aimed at refining the normals' accuracy. However, since the second stage is mainly aimed at fine-tuning, NGLO may face issues with normal flipping if the initial stage fails to achieve globally consistent normals.
SHS-Net is an end-to-end network that extracts both local, dense features and global, sparse features from the input point cloud. It orients points by penalizing discrepancies between these features in the latent space. Utilizing both local and global features is a double-edged sword. On one hand, it renders SHS-Net highly effective for simple models, where these features can be well aligned. On the other hand, it struggles with input models characterized by complex geometry and topology, primarily due to the significant discrepancies between local and global features. In Figure~\ref{fig:NGLO_SHSNET}, we compare our method to NGLO and SHS-Net for four models featuring thin structures and complex topology. To assess the robustness of these methods, we intentionally introduce noise or outliers to each test model. We observed that both NGLO and SHS-Net reconstruct the chick body well, which has a relatively simple geometry. However, they encounter difficulties with the thin structures of the chick's feet and fail to capture the complex topology of the other three models.

\paragraph{Limitations} Despite the demonstrated robustness and accuracy of our method, it is important to acknowledge its current limitations. As highlighted in  \cite{TakayamaJKS14}, the Dirichlet energy of GWN only measures the magnitude of the oscillation of a function and does not explicitly indicate the inside-outside orientation on non-manifold surfaces. As the proposed boundary energy is derived from the Dirichlet energy, our method is suitable only for point clouds sampled from manifold surfaces. Another limitation is the scalability issue due to the quadratic time complexity of our method. While iPSR~\cite{hou2022iterative} can handle large-scale models with millions of points, our method is constrained to smaller-scale models, with around 100K points. 

\section{Conclusions}
We introduce a new method for globally consistent orientation for point clouds. Our method initializes each point with a randomly assigned normal, and then iteratively updates its normal by maximizing the boundary energy of the generalized winding number field associated with the normals. When the algorithm terminates, we restore the harmonicity of the GWN field, whose gradients are the globally consistent orientations. Computational results show that our method is particularly effective in handling models with complex topology and thin structures, areas that existing methods have trouble with. 

\begin{acks} 
We thank the reviewers for their detailed comments and constructive suggestions. Special thanks are also due to Shepherd for his/her valuable feedback, which has significantly enhanced the clarity and quality of the paper. Additionally, we are thankful to Nicholas Sharp for his open-sourced visualization software Polyscope, which facilitated the creation of the images presented in the paper. This project was partially supported by the 
Beijing Municipal Science and Technology Commission and Zhongguancun Science Park Management Committee (No.Z221100002722020), the National Nature Science Foundation of China (No.62072045), the Nature Science Foundation of Beijing (No.7242167), the Ministry of Education, Singapore, under its Academic Research Fund Grants (MOE-T2EP20220-0005 \& RT19/22), and the RIE2020 Industry Alignment Fund–Industry Collaboration Projects (IAF-ICP) Funding Initiative, as well as cash and in-kind contribution from the industry partner(s).
\end{acks}

\newpage
\bibliographystyle{ACM-Reference-Format}
\bibliography{sample-base}
\newpage
\appendix

\begin{figure*}
  \centering
  
  \includegraphics[width=0.92\textwidth]{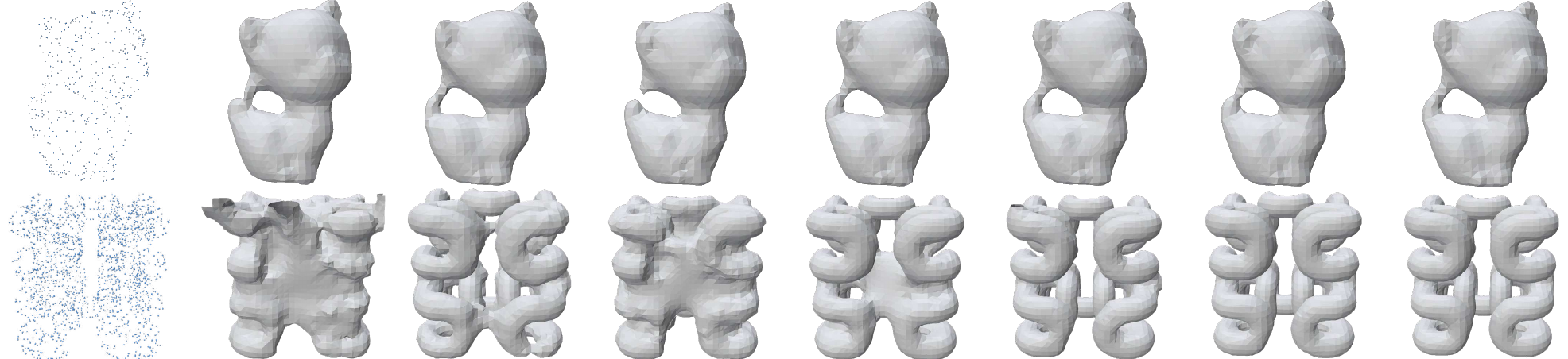}
  \includegraphics[width=0.92\textwidth]{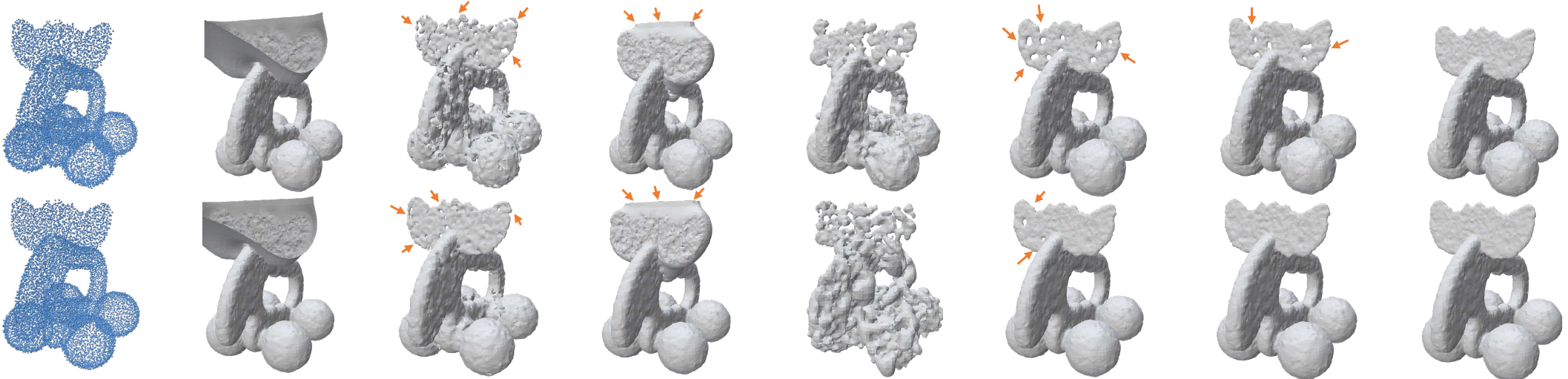}
  \includegraphics[width=0.92\textwidth]{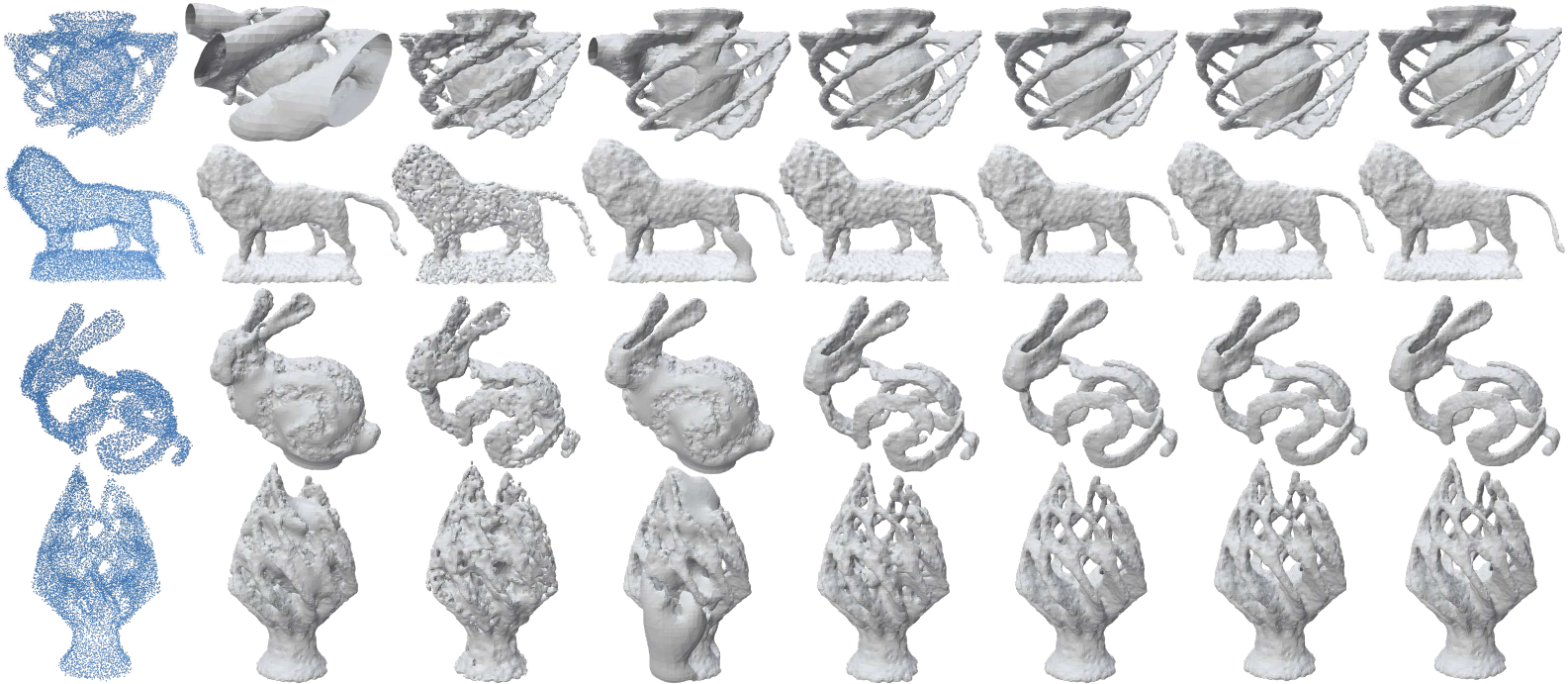}
   \includegraphics[width=1.0\textwidth]{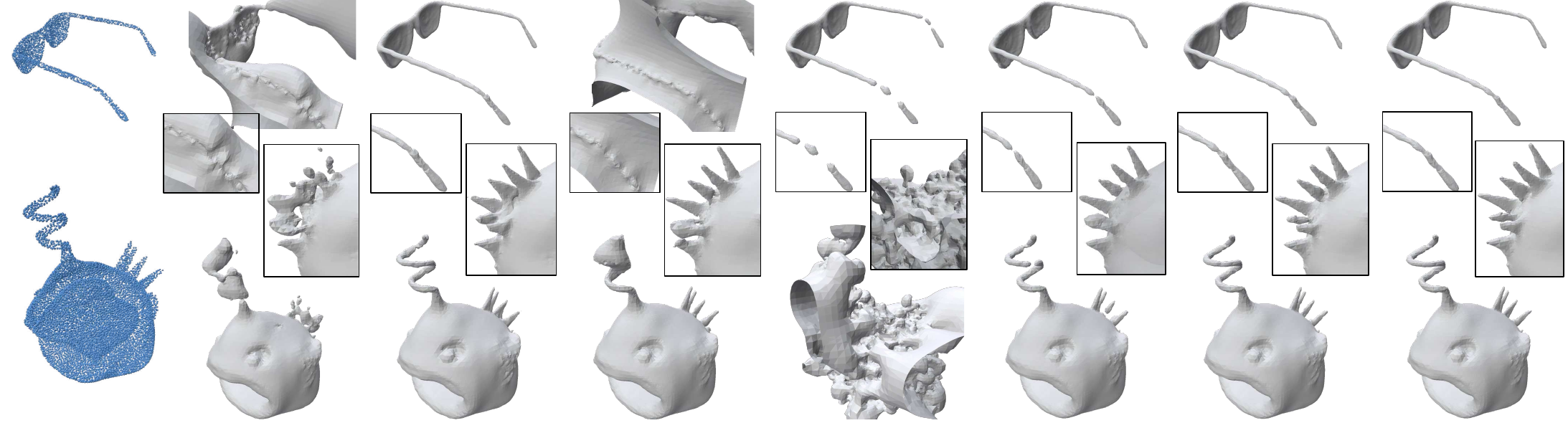}
   \put(-490,-20){\makebox(0,0)[bl]{ \textsf{Input}}}
   \put(-440,-20){\makebox(0,0)[bl]{ \textsf{Dipole}}}
   \put(-370,-20){\makebox(0,0)[bl]{ \textsf{PGR}}}
   \put(-310,-20){\makebox(0,0)[bl]{ \textsf{NeuralGF}}}
   \put(-230,-20){\makebox(0,0)[bl]{ \textsf{GCNO}}}
   \put(-170,-20){\makebox(0,0)[bl]{ \textsf{iPSR}}}
   \put(-110,-20){\makebox(0,0)[bl]{ \textsf{Ours}}}
   \put(-50,-20){\makebox(0,0)[bl]{ \textsf{Reference}}}
  \caption{Comparison with existing methods. Rows 1 \& 2: Sparse point clouds (Kitten: 500 points; Hilbert Cube: 2,500 points). Row 3 \& 4: The Elk model featuring thin structures with 0.5\% and 0.75\% noise level. Rows 5-8: Models with 0.75\% noisy level. Rows 9 \& 10: Two challenging models featuring thin plates, slender tubes, and closely adjacent faces.} 
  \label{fig:sparse_compare}
\end{figure*}

\begin{figure*}
  \includegraphics[width=0.9\textwidth]{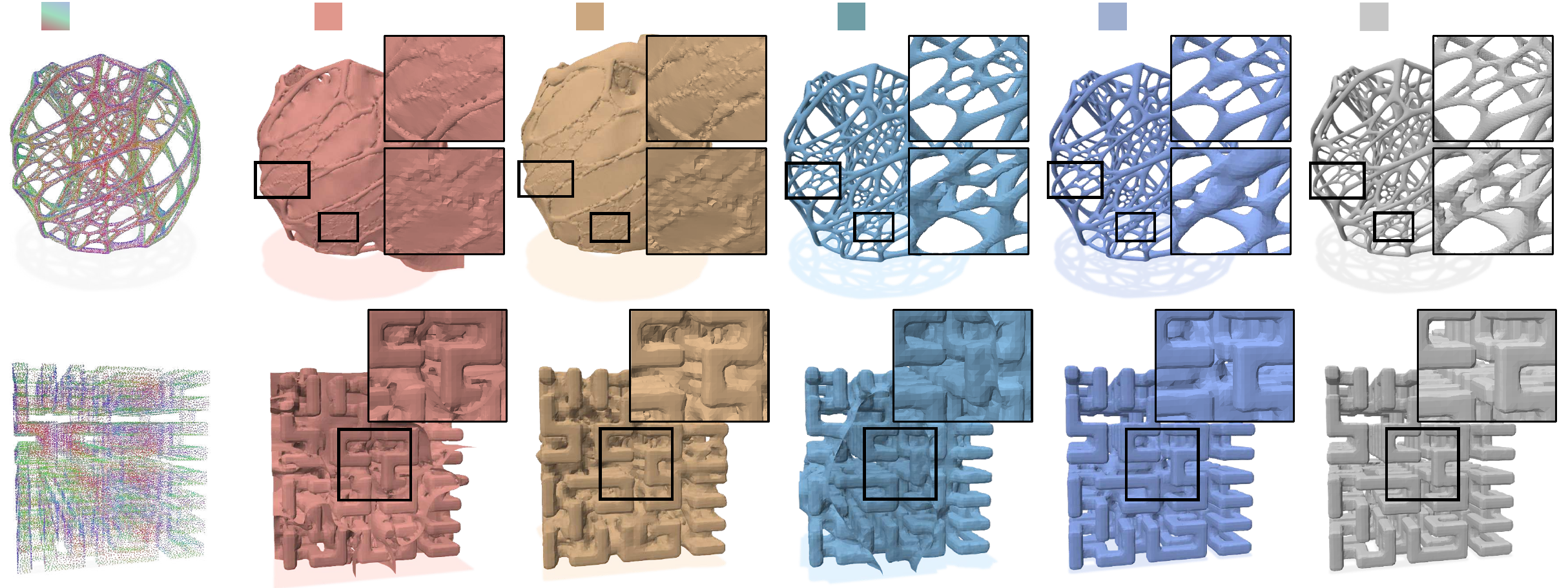}
  \centering
  \put(-435,163){\makebox(0,0)[bl]{ \textsf{Point Cloud}}}
  \put(-355,161){\makebox(0,0)[bl]{ \textsf{Dipole }}}
  \put(-279,163){\makebox(0,0)[bl]{ \textsf{NeuralGF }}}
  \put(-203,163){\makebox(0,0)[bl]{ \textsf{GCNO }}}
  \put(-127,163){\makebox(0,0)[bl]{ \textsf{iPSR }}}
  \put(-50,163){\makebox(0,0)[bl]{ \textsf{Ours }}}
  \put(-390,87){\makebox(0,0)[bl]{\small \textsf{89.4/58.1/6.750}}}
  \put(-310,87){\makebox(0,0)[bl]{\small \textsf{89.6/57.6/11.443}}}
  \put(-230,87){\makebox(0,0)[bl]{\small \textsf{7.90/13.2/0.228}}}
  \put(-150,87){\makebox(0,0)[bl]{\small \textsf{5.78/9.92/0.205}}}
  \put(-82,87){\makebox(0,0)[bl]{\small \textsf{\textbf{3.44}/\textbf{5.14}/\textbf{0.197} }}}
  \put(-390,-4){\makebox(0,0)[bl]{\small \textsf{88.3/72.9/6.510 }}}
  \put(-310,-4){\makebox(0,0)[bl]{\small \textsf{71.5/47.9/5.615
}}}
  \put(-230,-4){\makebox(0,0)[bl]{\small \textsf{85.6/67.1/6.222}}}
  \put(-150,-4){\makebox(0,0)[bl]{\small \textsf{28.7/38.1/1.398
}}}
  \put(-82,-4){\makebox(0,0)[bl]{\small \textsf{\textbf{19.4}/\textbf{20.5}/\textbf{1.257}}}}
  \caption{The reconstruction results of several state-of-the-art point orientation methods (Dipole \cite{metzer2021orienting}, NeuralGF \cite{li2023neuralgf}, GCNO \cite{xu2023globally}, iPSR \cite{hou2022iterative}) and our method.   
  Our method is more effective than the other methods at handling point clouds with complex geometries and topologies. The quantitative numbers below each model indicate the mean and the standard deviation of angle differences (degree), and Chamfer distances ($10^{-3}$), respectively.
  }
  \label{fig:teaser}
\end{figure*}

\begin{figure*}[!htbp]
  \centering  \includegraphics[width=1.0\linewidth]{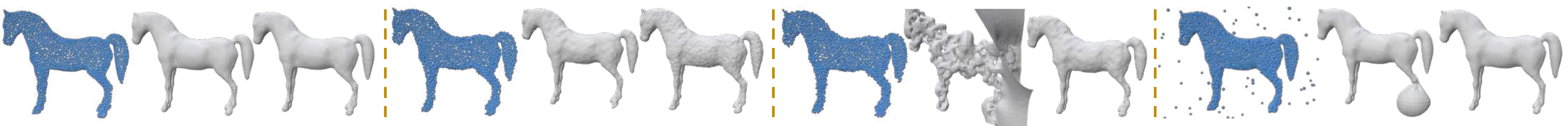}
  
  \vspace{-0.2in}
    \caption{Comparison with VIPSS on the horse model with 4k points under 3 different noise levels: 0.0\%, 0.5\% and 0.75\%. The last scenario is the results with 1\% outliers but no noise. For each scenario, we show the input point cloud, the results from VIPSS, and our results. While VIPSS handles clean data and low-level noise effectively, it struggles with high-level noise and outliers.} \label{fig:extension}
\end{figure*}

\begin{figure*}[!htbp]
  \centering
    \includegraphics[scale=0.5]{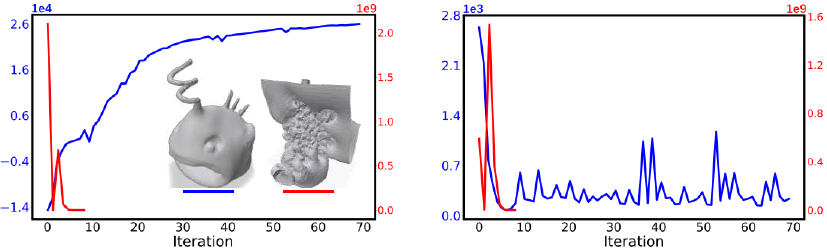}
      \put(-180,60){\small \makebox(0,0)[bl]{\textsf{Objective function}}}
    \put(-70,60){\small \makebox(0,0)[bl]{\textsf{Gradient norm}}}
    \put(-60,-4){\small \makebox(0,0)[bl]{\textsf{Iteration}}}
    \put(-160,-4){\small \makebox(0,0)[bl]{\textsf{Iteration}}}
  \includegraphics[width=4.3in]{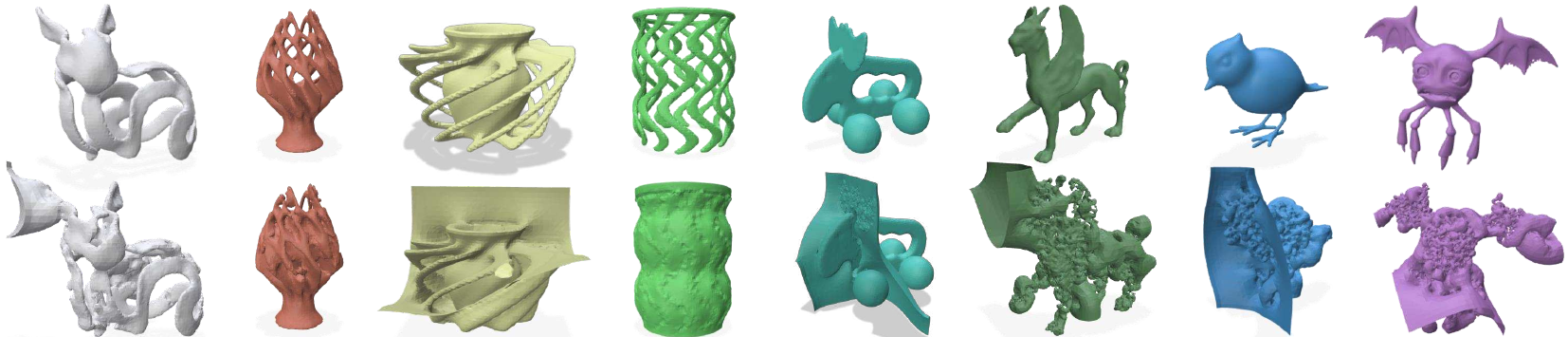}
  \caption{The efficacy of GCNO heavily relies on a uniform distribution of Voronoi vertices both inside and outside the point cloud. This condition is often unattainable for models characterized by complex geometry and topology, leading to inaccuracies in point orientation and, consequently, subpar reconstruction outcomes. In the left, we plot the the objective functions and gradient norms for GCNO (red) and our method (blue) for an example where GCNO fails due to stopping optimizing its objective at an early stage. In the right, we show our results (top) and GCNO's (bottom). }
  \label{fig:GCNO_all_fail}
\end{figure*}

\begin{figure*}
  \setlength{\unitlength}{1mm}
\includegraphics[width=1\linewidth]{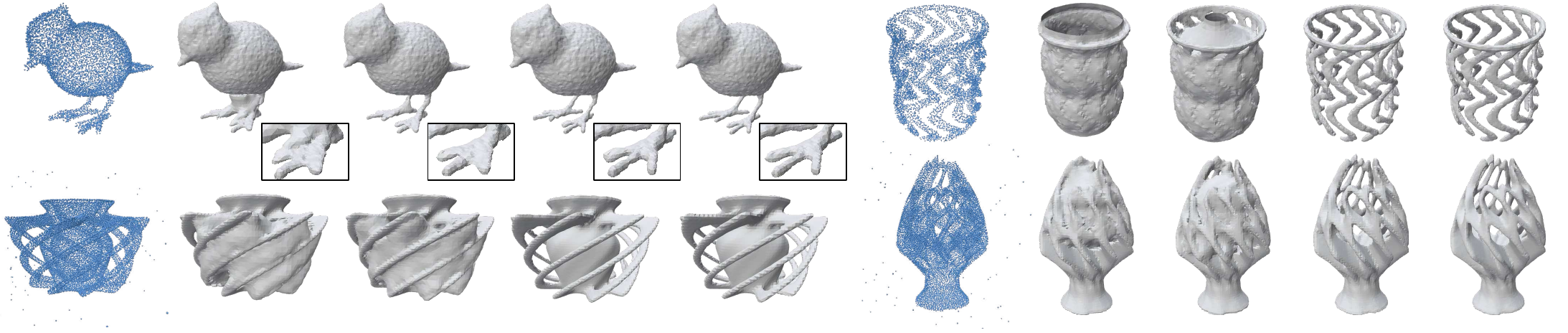}
\put(-76, 1){\makebox(0,0)[tl]{ \textsf{Input}}}
\put(-60, 1){\makebox(0,0)[tl]{ \textsf{NGLO}}}
\put(-45, 1){\makebox(0,0)[tl]{ \textsf{SHS-Net}}}
\put(-27, 1){\makebox(0,0)[tl]{ \textsf{Ours}}}
\put(-15, 1){\makebox(0,0)[tl]{ \textsf{Reference}}}
\put(-175, 1){\makebox(0,0)[tl]{ \textsf{Input}}}
\put(-155, 1){\makebox(0,0)[tl]{ \textsf{NGLO}}}
\put(-138, 1){\makebox(0,0)[tl]{ \textsf{SHS-Net}}}
\put(-117, 1){\makebox(0,0)[tl]{ \textsf{Ours}}}
\put(-100, 1){\makebox(0,0)[tl]{ \textsf{Reference}}}
\put(-148, 28){\makebox(0,0)[tl]{\tiny \textsf{(12.3,32.3,0.43)}}} %
\put(-130, 28){\makebox(0,0)[tl]{\tiny \textsf{(8.39,20.6,0.10)}}} %
\put(-110, 28){\makebox(0,0)[tl]{\tiny \textsf{(11.8,10.9,0.09)}}} %
\put(-52, 22){\makebox(0,0)[tl]{\tiny \textsf{(83.8,70.2,4.68)}}} %
\put(-37, 22){\makebox(0,0)[tl]{\tiny \textsf{(72.5,68.2,4.71)}}} %
\put(-22, 22){\makebox(0,0)[tl]{\tiny \textsf{(26.7,18.4,0.24)}}} %
\put(-148, 4){\makebox(0,0)[tl]{\tiny \textsf{(76.8,83.6,4.31)}}} %
\put(-130, 4){\makebox(0,0)[tl]{\tiny \textsf{(55.3,76.2,4.26)}}} %
\put(-110, 4){\makebox(0,0)[tl]{\tiny \textsf{(8.30,8.55,0.28)}}} %
\put(-52, 4){\makebox(0,0)[tl]{\tiny \textsf{(48.6,67.0,26.0)}}} %
\put(-36, 4){\makebox(0,0)[tl]{\tiny \textsf{(53.7,68.9,26.8)}}} %
\put(-21, 4){\makebox(0,0)[tl]{\tiny \textsf{(16.7,21.3,0.18)}}} %
\caption{Comparison with the SOTA deep-learning approaches on point clouds subjected to 0.75\% noise levels (top) and 1\% of outliers (bottom). The 3-tuple in each figure is the error metrics, $\mu$ (degree), $\sigma$ (degree) and CD ($10^{-3}$).}
 \label{fig:NGLO_SHSNET}
\end{figure*}
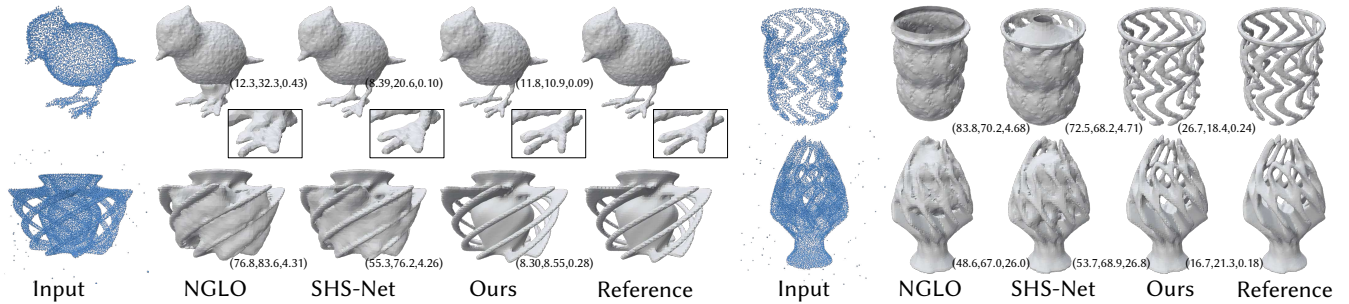

\end{document}


\newcommand{\lwzb}[1]{\textcolor{orange}{#1}}

\title{Consistent Point Orientation for Manifold Surfaces via Boundary Integration (Supplementary Material) }

\maketitle

\section{Implementation Details}

\paragraph{Discretizing function $g$.} We define the function
$$g\left(w(\mathbf{p}_i^{\pm})\right)=\left(1-e^{\max\{w(\mathbf{p}_i^{+}), 1\}}\right) +\left(1- e^{\max\{w(\mathbf{p}_i^{-}), 1\}}\right),$$ to penalize values of $w$ outside the interval $[0,1]$. Given the numerical errors introduced by discretization, the ideal range of $w$ might not strictly be $[0,1]$. Consequently, we relax the constraint from $\max\{w(\mathbf{p}_i^{\pm}), 1\}$ to $\max\{w(\mathbf{p}_i^\pm), 1 + \delta\} $. In our experiments, we empirically set $\delta = 0.05$ for all test models.

\paragraph{Computing $a_i$.} 
The geodesic Voronoi area $a_i$ serves as a weight, reflecting the contribution of point $\mathbf{p}_i$ to the surface integral. Computing geodesic Voronoi diagrams on point clouds is computationally intensive, thus a common simplification involves using 2D Voronoi areas. For oriented point clouds, where the normal at each point is  available,  \citet{Barill2018FastWN} computed $a_i$ by projecting $\mathbf{p}_i$ and its $k$-nearest neighbors onto the tangent plane at $\mathbf{p}_i$.
Since our input point cloud $\mathcal{P}$ is unoriented, we first fit a tangent plane using $\mathbf{p}_i$ and its $k$-nearest neighbors, then proceed as  \citet{Barill2018FastWN} did. We empirically set the neighborhood size to 15 in our implementation. 

\section{Boundary Integration}
Observing the time-consuming nature of computing the Dirichlet energy via  volume integral, 
\citet{TakayamaJKS14} employed Green's first identity to transform volume integral into boundary integrals. In this section, we provide a brief overview of the derivation by \citet{TakayamaJKS14} and refer readers to their paper for more technical details.  

Consider the generalized winding number $w$ induced by consistently oriented normals $\hat{\mathbf{n}}$,  $$w(\mathbf{x})=\int_{\partial \Omega}\frac{\partial G(\mathbf{x}, \mathbf{z})}{\partial \hat{\mathbf{n}}_\mathbf{z}}\mathrm{d}\mathbf{z}.$$

The Dirichlet energy $\int_{\mathbb{R}^3 \setminus \partial \Omega}\|\nabla w\|^2$ can be decomposed into two separate integrals for interior and exterior regions, $$\int_{\mathbb{R}^3 \setminus \partial \Omega}\|\nabla w\|^2 = \int_{\Omega^+}\|\nabla w\|^2 + \int_{\Omega^-}\|\nabla w\|^2.$$

Denote by $w^+$ and $w^-$ the GWN on $\partial\Omega$ through the limits of approaching $\partial\Omega$ from the outward and inward normal directions, respectively. The values $w^+$ and $w^-$ represent the jump boundary values of Poisson's equation.

Notice that the directional derivatives $\nabla_{\hat{\mathbf{n}}} w$ are vector dot products $\langle \nabla w, {\hat{\mathbf{n}}}\rangle$. 
Applying Green's first identity\footnote{Although the exterior domain $\Omega^+$ is not compact, the first Green's identity remains applicable to the Dirichlet energy $\int_{\Omega^+}\|\nabla w\|^2$ due to the characteristics of the Poisson kernel. Given that the Poisson kernel is twice continuously differentiable $C^2$ in $\Omega^+$ and satisfies $\Delta P = 0$, it is $L_2$-integrable over $\Omega^+$.} to $\int_{\Omega^+}\|\nabla w\|^2$ yields
\begin{align}
\int_{\Omega^+}\|\nabla w\|^2 &= \int_{\Omega^+}\langle \nabla w, \nabla w\rangle \nonumber \\
   &= -\int_{\Omega^+}w\Delta w + \int_{\partial \Omega^+}w^+\nabla_{\hat{n}^+} w^+ \nonumber \\
   &=\int_{\partial \Omega^+}w^+\nabla_{\hat{n}^+} w^+. \nonumber 
\end{align}  
The integral $\int_{\Omega^+}w\Delta w$ vanishes since the GWN is harmonic for points off the surface, i.e., $\Delta w(\mathbf{x}) = 0$ for $\mathbf{x} \in \mathbb{R}^3 \setminus \partial\Omega$.

Due to symmetry in the treatment of the interior and exterior regions, we can obtain the integral for the interior region $\Omega^-$ in a similar fashion.

Due to the Neumann boundary condition, which states
$$
\frac{\partial u^{+}(\mathbf{x})}{\partial \mathbf{n}_{\mathbf{x}}} =  \frac{\partial u^{-}(\mathbf{x})} {\partial \mathbf{n}_{\mathbf{x}}}, \;\;\mathbf{x} \in\; \partial \Omega, 
$$
we can conclude that the gradients of $w$ along the normal directions on the boundary are equal, leading to  $$\nabla_{\hat{\mathbf{n}}^{-}_{\mathbf{x}}} w^-(\mathbf{x}) = \nabla_{\hat{\mathbf{n}}^{-}_{\mathbf{x}}} w^+(\mathbf{x})$$
for boundary points $\mathbf{x}\in\partial\Omega$.
Additionally, with the jump boundary condition of Poisson's equation $w^+-w^-$ being a constant, we have $\nabla w^-(\mathbf{x}) = \nabla w^+(\mathbf{x})$ for $\mathbf{x}\in\partial\Omega$.

Since $\hat{\mathbf{n}}^+=-\hat{\mathbf{n}}^-$, the Dirichlet energy becomes an integral over the boundary surface \cite{TakayamaJKS14}:
\begin{equation}
    \int_{\mathbb{R}^3 \setminus \partial \Omega}\|\nabla w\|^2=\int_{\partial \Omega}(w^- - w^+)\nabla_{\hat{\mathbf{n}}} w^{-}.
    \label{eqn:takayama}
\end{equation}
Equation (\ref{eqn:takayama}) motivates us to define the boundary energy by treating normals $\bf n$ as variables. Note that the boundary energy and the Dirichlet energy coincide only when the normals are globally consistent, i.e.,  $\mathbf{n}=\hat{\mathbf{n}}$.

\section{Comparison with GCNO for Scalability}
\begin{figure}[htbp]
  \begin{minipage}[t]{0.5\linewidth}
    \vspace{-0.5in}
    \begin{tabular}{l|lllll}
    \hline
    \# points & 10k & 20k & 40k & 80k & 100k \\ \hline
    GCNO  & 66   & 188   & DNF   & DNF   & DNF    \\ \hline
    Ours   & 17   & 48   & 187   & 429   & 1100    \\ \hline
    \end{tabular}
    \label{table:compare_with_GCNO}
  \end{minipage}%
  \begin{minipage}[t]{0.5\linewidth} %
    \hspace{0.8in}
    \includegraphics[width=0.3\linewidth]{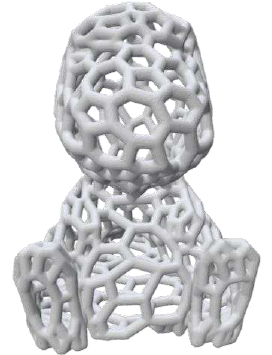}
    \end{minipage}

    \caption{Scalability test conducted on the Dino model across five different resolutions, reporting the running time in minutes. ``DNF'' denotes that GCNO did not finish within a 24-hour period.}
    \label{fig:compare_with_GCNO}
\end{figure}

We conduct comparative experiments with GCNO on a model featuring thin structures and complex topology. We sample the model at different five resolutions: 10k, 20k, 40k, 80k, and 100k points. The runtime for both our method and GCNO is presented in Figure \ref{fig:compare_with_GCNO}. We observe that GCNO performs optimally for models with up to 10k points, typically generating results within an hour. However, for larger models containing 40k points, GCNO is unable to complete within a 24-hour period. In contrast, our method can produce  results in just three hours.

\section{Additional Results}
\begin{figure}
  \centering  
  \includegraphics[width=\linewidth]{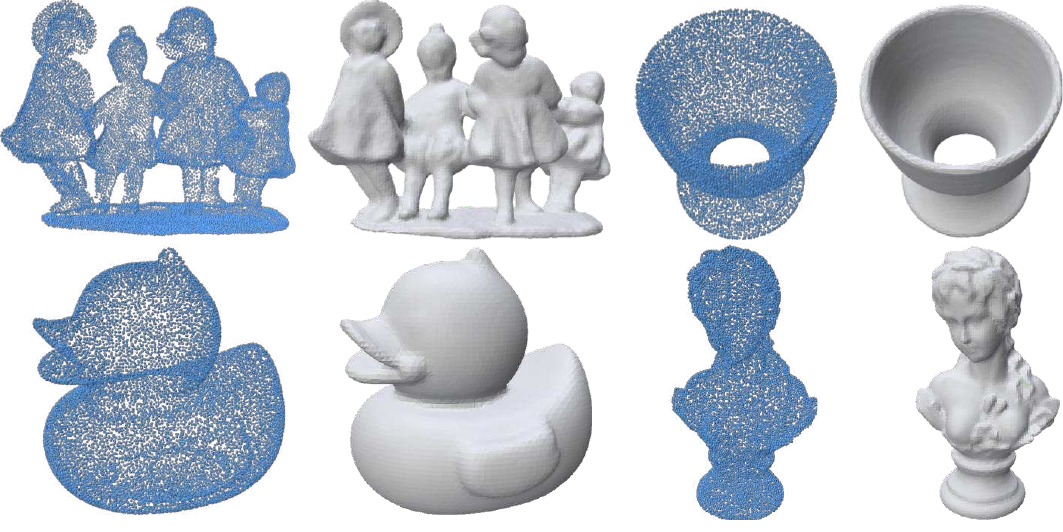}
  \includegraphics[width=\linewidth]{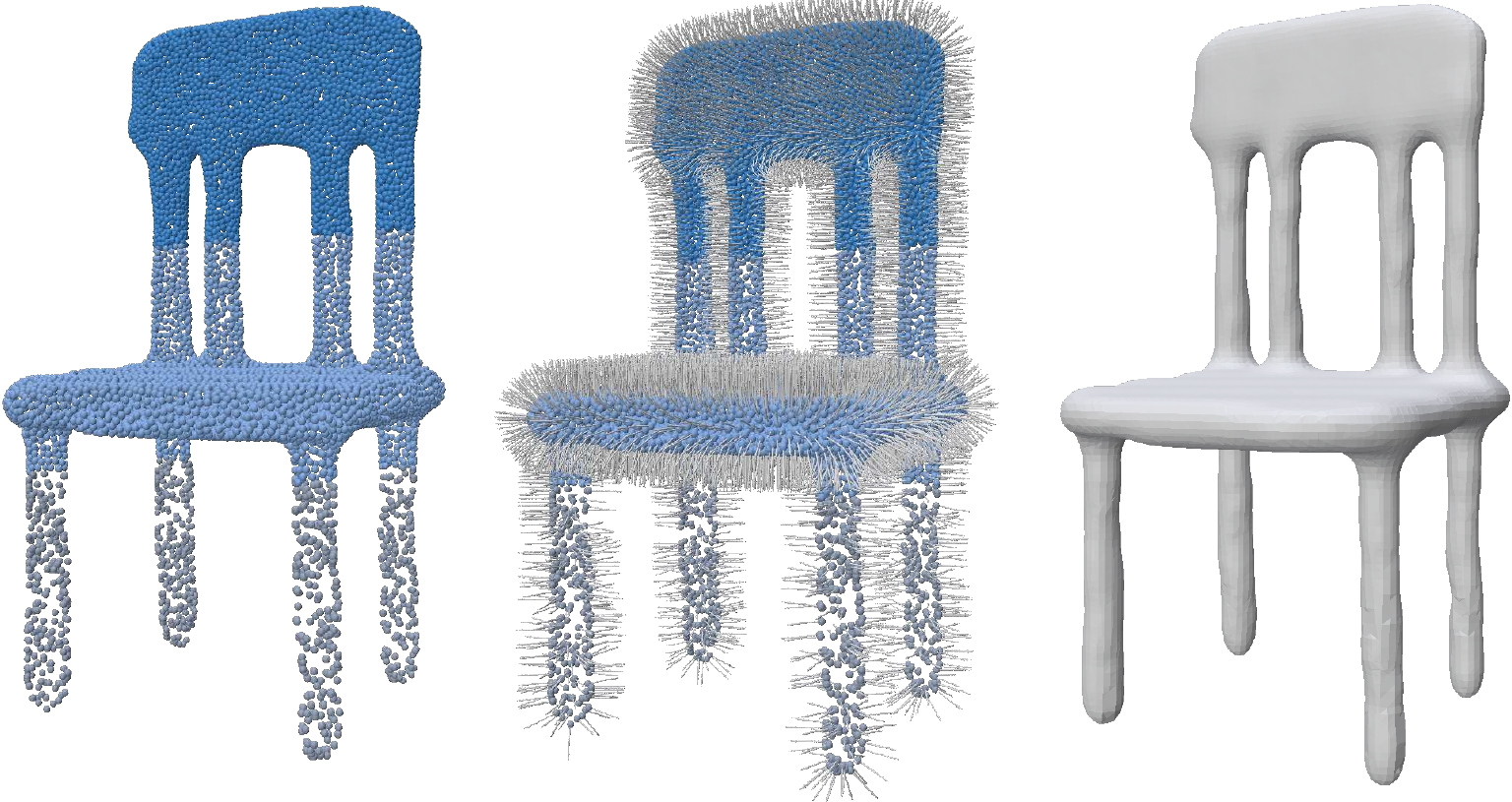}
  \caption{Additional results. Top \& middle: Four real scan point clouds from \cite{huang2022surface} and \cite{DeepGeo}. We downsampled them to between 10k and 20k points each. Bottom: A model with varying sampling densities. Different colors represent different densities. }
  \label{fig:scan}
\end{figure}

\begin{figure}
    \centering
    \includegraphics[width=0.8\linewidth]{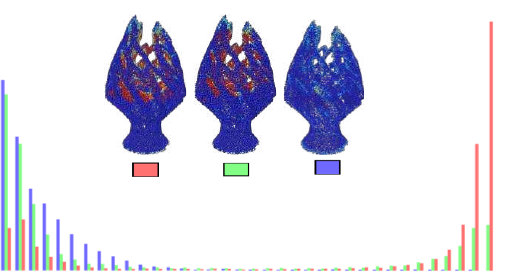}
    \put(-150, 35){\makebox(0,0)[tl]{\small \textsf{NGLO}}}
    \put(-120, 35){\makebox(0,0)[tl]{\small \textsf{SHS-Net}}}
    \put(-78, 35){\makebox(0,0)[tl]{\small \textsf{Ours}}}
    \put(-200, 0){\makebox(0,0)[tl]{ \textsf{0}}}
    \put(-10, 0){\makebox(0,0)[tl]{ \textsf{180}}}
    \put(-215, 20){\makebox(0,0)[tl]{ \textsf{1000}}}
    \put(-215, 40){\makebox(0,0)[tl]{ \textsf{2000}}}
    \put(-215, 60){\makebox(0,0)[tl]{ \textsf{3000}}}
    \put(-215, 80){\makebox(0,0)[tl]{ \textsf{4000}}}
    \put(-130, -1){\makebox(0,0)[tl]{ \textsf{Angle difference}}}
    \put(-215, 40){\makebox(0,0)[r]{\rotatebox{90}{\textsf{points}}}}
    \caption{Distributions of angular discrepancies among our method and two state-of-the-art deep learning approaches, NGLO~\cite{li2023neural} and SHS-Net~\cite{li2023shsnet}. We visualize the color-encoded angle difference and present the corresponding distribution, with angles spanning from $0^\circ$ to $180^\circ$. Due to the lack of global consistency in their predicted normals, utilizing them in Poisson surface reconstruction leads to suboptimal reconstruction results.}
    \label{fig:AA_distribution}
\end{figure}

Figure~\ref{fig:scan} shows additional visual results and   Figure~\ref{fig:AA_distribution} illustrates the distribution of angular differences among our method and two SOTA deep learning methods.
Table~\ref{tab:table_0.5} presents the experimental results under noise level of 0.5\% and 0.75\%, respectively.

\begin{figure*}
  \centering
\includegraphics[width=0.85\linewidth]{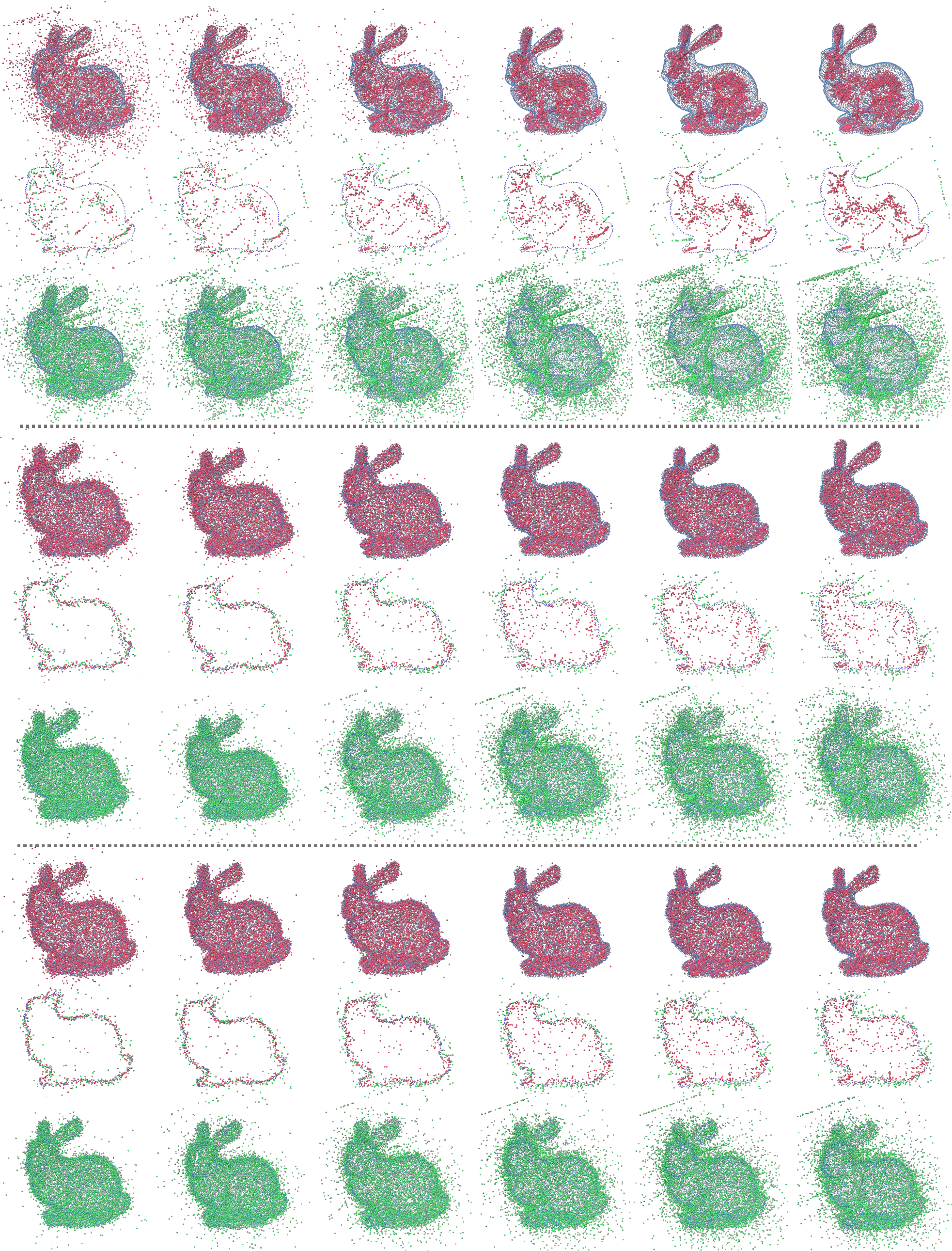}
    \put(-420,-2.5){\makebox(0,0)[bl]{Initialization}}
    \put(-335,-2){\makebox(0,0)[bl]{5 iter}}
    \put(-265,-2.5){\makebox(0,0)[bl]{10 iter}}
    \put(-190,-2){\makebox(0,0)[bl]{15 iter}}
    \put(-120,-2){\makebox(0,0)[bl]{20 iter}}
    \put(-40,-3){\makebox(0,0)[bl]{Final}}
  \caption{Visualization of $\mathbf{p}^\pm$ during the optimization process for the Bunny model under noise levels of 0\% (top), 0.5\% (middle), and 0.75\% (bottom). Red points represent exterior samples $\mathbf{p}^+$, while green points represent interior samples $\mathbf{p}^-$. For better visualization, cut views are provided. Initially, some points $\mathbf{p}_i$ exhibit inside-out configurations, where the green samples $\mathbf{p}_i^+$ are inside the surface and the red samples $\mathbf{p}_i^-$ outside. The frequency of such configurations decreases rapidly during the boundary energy optimization process.}  
  \label{fig:in_and_out}
\end{figure*}

In Figure \ref{fig:in_and_out}, we visualize the distribution of $\mathbf{p}^\pm$ during the optimization process for the Bunny model at three noise levels: 0\%, 0.5\%, and 0.75\%. Initially, some points exhibit an inside-out configuration, where $\mathbf{p}^+$ is positioned inside and $\mathbf{p}^-$ outside. The frequency of such configurations decreases rapidly. Additionally, for this model, we have never observed instances where $\mathbf{p}^+$ and $\mathbf{p}^-$ end up on the same side.

\newpage
\bibliographystyle{ACM-Reference-Format}
\bibliography{sample-base}

\newpage

\begin{table*}
  \caption{The 5 values are  mean $\mu$ and standard deviation of angle differences (degree),  Chamfer distance ($10^{-3}$), peak memory consumption (MB) and  running time (s). 
  }
  \label{tab:table_0.5}
  \centering
  \resizebox{\textwidth}{!}{
  \fontsize{10}{10}\selectfont
  \begin{tabular}{c|cccccc}
    \toprule
    Sampling & \multicolumn{6}{c}{Sampling with 0.50\% noise}\\
    \midrule
    Models 
    & Dipole & PGR & iPSR & NeuralGF & GCNO & Ours\\
    \midrule
    
    30cup (11311)
    & 6.4/17/\textbf{0.12}/65/2.5 & 67/82/0.17/\textbf{13}/1.1 & \textbf{5.5}/9.8/\textbf{0.12}/16/0.13 & 6.1/7.8/\textbf{0.12}/640/1.9 & 13/170/0.13/$>10^4$/0.18 & 5.6/\textbf{7.2/0.12}/770/\textbf{0.096}\\
    
    397horse (8784) 
    & 15/39/1.2/50/2.4 & 74/90/0.67/\textbf{9.3}/0.86 & \textbf{6.4}/8.6/\textbf{0.076}/16/0.10 & 11/20/0.37/640/1.9 & 16/19/0.084/6000/0.14 & 7.4/\textbf{9.6/0.076}/460/\textbf{0.086 }\\
    
    Lion (13138)  
    & 15/28/0.38/71/3.2 & 80/95/1.1/17/1.4 & \textbf{12/16/0.097}/18/0.11 & 15/19/0.081/640/1.9 & 22/28/0.084/$>10^4$/0.20 & 12/17/0.073/1000/\textbf{0.087}\\

    Bird (8187)  
    & 13/37/3.4/50/2.0 & 67/82/0.24/\textbf{8.6}/0.80 & \textbf{4.9/7.7/0.075}/14/0.10 & 7.7/9.9/0.073/640/1.9 & 16/26/0.085/5400/0.13 & 5.8/9.2/0.071/400/\textbf{0.084}\\

    Botijo (11305)
    & 25/58/9.1/64/2.4 & 72/86/0.49/\textbf{11}/1.1 & \textbf{5.6/7.6/0.088}/15/0.13 & 17/37/1.7/640/1.9 & 12/14/0.088/$>10^4$/0.17 & 7.0/9.0/0.087/760/\textbf{0.086}\\

    BunnyPeel (8648) 
    & 85/110/10/53/1.9 & 70/81/0.66/\textbf{10}/0.85 & 17/25/0.14/21/0.11 & 77/100/12/640/1.9 & 16/22/0.15/6600/0.14 & \textbf{16/21/0.14}/450/\textbf{0.092}\\

    Candle (12221)
    & 64/92/10/67/2.4 & 71/81/0.66/\textbf{20}/1.2 & 20/27/0.14/37/0.14 & 28/43/12/640/1.9 & 36/51/0.15/$>10^4$/0.19 & \textbf{18/24/0.14}/910/\textbf{0.090}\\

    Elk (12541)
    & 17/46/8.4/77/2.2 & 71/85/0.94/\textbf{21}/1.3 & \textbf{5.9/10/0.12/21}/0.12 & 22/51/12/640/1.9 & 86/95/2.6/$>10^4$/0.19 & 7.6/11/0.12/1200/\textbf{0.10}\\

    Felinei (15057)
    & 29/60/9.1/81/3.0 & 78/92/0.70/\textbf{24}/1.7 & \textbf{11/18/0.12}/26/0.12 & 33/60/4.6/640/1.9 & 88/95/2.2/$>10^4$/0.23 & 12/17/0.10/1400/\textbf{0.093}\\

    Horse (8657)
    & 10/23/0.12/50/2.4 & 74/90/1.56/\textbf{9.3}/0.86 & \textbf{6.4/86/0.086}/17/0.10 & 9.9/13/0.089/640/1.9 & 16/19/0.090/5800/0.13 & 7.4/9.6/0.085/450/\textbf{0.086}\\

    linkCupTop (8602)
    & 90/110/4.7/58/2.1 & 51/60/0.35/\textbf{7.7}/0.85 & 29/35/0.28/71/0.11 & 87/110/4.6/640/1.9 & \textbf{18/22/0.20}/5200/0.14 & 18/22/0.22/450/\textbf{0.095}\\

    Pulley (16116)
    & 15/31/0.14/86/3.1 & 75/90/0.30/26/1.8 & \textbf{9.1/12/0.092/15}/0.16 & 16/23/0.11/640/1.9 & 20/24/0.11/$>10^4$/0.24 & 10/13/0.087/1600/\textbf{0.093}\\

    Vase (13375)
    & 53/87/10/81/2.6 & 60/70/0.68/\textbf{18}/1.4 & 13/18/0.25/49/0.14 & 71/98/4.6/640/1.9 & 14/22/0.28/$>10^4$/0.21 & \textbf{12/15/0.26}/1100/\textbf{0.10}\\

    BS (4000)
    & 4.9/6.1/0.13/28/2.1 & 41/48/0.14/\textbf{2.5}/0.52 & \textbf{4.2/5.0/0.14}/9.8/0.094 & 6.4/7.4/0.15/640/1.9 & 6.0/7.1/0.17/2000/\textbf{0.067} & 4.4/5.3/0.13/97/0.075\\

    Bunny (8830)
    & 16/41/15/61/2.1 & 65/79/0.32/\textbf{7.9}/0.86 & \textbf{6.4/8.5/0.15}/13/0.11 & 7.9/9.4/0.17/640/1.9 & 12/14/0.14/6400/0.14 & 7.1/9.0/0.13/470/\textbf{0.094}\\

    CupPossion (10130)
    & 6.0/22/0.18/60/2.2 & 55/69/0.15/\textbf{9.3}/1.0 & \textbf{3.0/3.9/0.19}/14/0.11 & 5.5/6.4/0.14/640/1.9 & 6.4/7.5/0.16/8300/0.16 & 3.6/4.7/0.14/630/\textbf{0.089}\\

    Fandisk (13854)
    & 6.8/11/0.16/78/2.3 & 75/91/0.16/24/1.5 & \textbf{6.5/9.5/0.14/19}/0.18 & 7.3/9.3/0.14/640/1.9 & 17/21/0.20/$>10^4$/0.21 & 6.9/9.3/0.14/1200/\textbf{0.093}\\

    Fertility (13802)
    & 7.9/21/0.098/74/3.2 & 75/91/0.64/\textbf{20}/1.5 & \textbf{5.4/7.0/0.087}/22/0.15 & 8.0/10/0.089/640/1.9 & 14/17/0.13/$>10^4$/0.21 & 5.8/7.5/0.082/1200/\textbf{0.089}\\

    Average 
    & 27/47/4.1/64/2.4 & 68/81/0.77/\textbf{14}/1.1 & 9.5/13/0.14/23/0.12 & 24/35/2.3/640/1.9 & 24/29/0.42/$>10^4$/0.17 & \textbf{9.3/12/0.12}/880/\textbf{0.091}\\

    \bottomrule
  \end{tabular}
  }
\end{table*}

\begin{table*}%
  \label{tab:table_0.75}
  \centering
  \resizebox{\textwidth}{!}{
  \fontsize{10}{10}\selectfont
  \begin{tabular}{c|cccccc}
    \toprule
    Sampling & \multicolumn{6}{c}{Sampling with 0.75\% noise}\\
    \midrule
    Models 
    & Dipole & PGR & iPSR & NeuralGF & GCNO & Ours\\
    \midrule
    
    30cup (11311)
    & 9.3/22/0.20/64/2.5 & 81/97/0.77/17/1.1 & 10/24/0.19/\textbf{15}/0.12 & 30/59/39/640/1.9 & 36/47/0.38/$>10^4$/0.19 & \textbf{7.6/9.6/0.15}/840/\textbf{0.096} \\
    
    397horse (8784) 
    & 15/36/0.62/50/2.4 & 83/98/0.87/\textbf{9.6}/0.86 & \textbf{8.4/11/0.091}/16/0.10 & 14/23/1.1/640/1.9 & 30/35/0.14/6400/0.14 & 10/13/0.092/500/\textbf{0.083}\\
    
    Lion (13138)  
    & 20/37/0.69/72/3.2 & 86/100/0.45/\textbf{18}/1.4 & \textbf{14/19/0.11}/21/0.12 & 20/31/1.4/640/1.9 & 34/41/0.12/$>10^4$/0.20 & 16/21/0.099/1100/\textbf{0.087}\\

    Bird (8187)  
    & 14/36/2.4/50/2.0 & 79/96/3.2/\textbf{9.1}/0.80 & \textbf{6.6/9.7/0.090}/18/0.11 & 10/14/0.093/640/1.9 & 24/30/0.095/5600/0.13 & 9.1/15/0.085/440/\textbf{0.085}\\

    Botijo (11305)
    & 26/58/7.6/62/2.4 & 82/97/6.7/\textbf{14}/1.1 & \textbf{7.5/9.9/0.098}/15/0.13 & 20/39/2.9/640/1.9 & 24/28/0.11/$>10^4$/0.17 & 9.6/12/0.098/830/\textbf{0.086}\\

    BunnyPeel (8648) 
    & 82/110/8.7/53/2.0 & 80/92/2.0/\textbf{8.4}/0.85 & 27/44/0.17/25/0.11 & 71/94/8.0/640/1.9 & 26/34/0.21/7000/0.14 & \textbf{23/29/0.18}/490/\textbf{0.092}\\

    Candle (12221)
    & 65/92/8.7/67/2.4 & 79/90/20/\textbf{15}/1.2 & \textbf{22/30/0.17}/50/0.13 & 61/85/8.0/640/1.9 & 46/60/0.21/$>10^4$/0.19 & 24/32/0.18/990/\textbf{0.090}\\

    Elk (12541)
    & 18/45/9.2/76/2.2 & 80/95/1.9/\textbf{22}/1.3 & \textbf{8.5/14/0.14}/37/0.13 & 23/50/9.6/640/1.9 & 69/80/0.72/$>10^4$/0.19 & 11/16/0.14/1300/\textbf{0.10}\\

    Felinei (15057)
    & 61/94/77/81/3.1 & 85/98/0.86/\textbf{22}/1.6 & 16/24/0.13/49/0.12 & 31/53/14/640/1.9 & 85/93/0.98/$>10^4$/0.23 & \textbf{15/21/0.12}/1500/\textbf{0.093}\\

    Horse (8657)
    & 14/30/0.24/50/2.4 & 83/98/0.83/\textbf{9.6}/0.86 & \textbf{8.4/24/0.10}/19/0.10 & 14/53/0.28/640/1.9 & 30/93/0.16/6300/0.14 & 10/21/0.11/490/\textbf{0.085}\\

    linkCupTop (8602)
    & 89/110/5.6/59/2.0 & 64/76/1.2/\textbf{8.4}/0.85 & 40/51/0.27/69/0.11 & 87/110/5.3/640/1.9 & 31/40/0.46/5700/0.14 & \textbf{26/32/0.24}/490/\textbf{0.096}\\

    Pulley (16116)
    & 17/31/0.13/85/3.1 & 83/98/0.44/28/1.8 & \textbf{13/17/0.12/17}/0.18 & 16/21/0.11/640/1.9 & 33/40/0.18/$>10^4$/0.24 & 14/17/0.098/1700/\textbf{0.093}\\

    Vase (13375)
    & 50/83/8.0/81/2.4 & 72/83/1.3/\textbf{20}/1.4 & \textbf{14/20/0.26}/51/0.14 & 38/59/3.5/640/1.9 & 20/30/0.44/$>10^4$/0.21 & 16/20/0.26/1200/\textbf{0.10}\\

    BS (4000)
    & 5.2/6.4/0.14/28/1.9 & 53/65/0.25/\textbf{2.7}/0.51 & \textbf{4.8/5.8/0.15}/10/0.094 & 7.6/8.7/0.13/640/1.9 & 7.8/9.0/0.21/2100/\textbf{0.067} & 5.2/6.1/0.13/110/0.075\\

    Bunny (8830)
    & 9.8/18/0.19/62/2.1 & 77/92/1.7/\textbf{9.4}/0.86 & \textbf{7.8/10/0.16}/14/0.11 & 9.9/12/0.16/640/1.9 & 22/27/0.16/7000/0.14 & 9.0/11/0.15/510/\textbf{0.094}\\

    CupPossion (10130)
    & 7.1/23/0.19/60/2.2 & 68/83/0.35/\textbf{11}/1.0 & \textbf{4.2/5.5/0.17}/13/0.11 & 7.1/8.3/0.22/640/1.9 & 14/17/0.22/8800/0.16 & 5.6/6.8/0.14/690/\textbf{0.089}\\

    Fandisk (13854)
    & \textbf{8.4/13/0.22}/79/2.3 & 83/99/1.2/24/1.5 & 8.7/12/0.20/\textbf{23}/0.18 & 9.6/12/0.20/640/1.9 & 30/36/0.18/$>10^4$/0.21 & 10/13/0.20/1300/\textbf{0.093}\\

    Fertility (13802)
    & 8.8/19/0.099/74/3.4 & 84/99/0.51/\textbf{21}/1.5 & \textbf{7.2/9.0/0.10}/28/0.15 & 9.9/12/0.10/640/1.9 & 27/32/0.13/$>10^4$/0.21 & 7.8/9.8/0.090/1300/\textbf{0.089}\\

    Average
    & 29/48/6.8/64/2.5 & 78/92/1.5/\textbf{15}/1.1 & \textbf{13}/18/0.15/27/0.12 & 27/40/5.0/640/1.9 & 33/40/0.34/$>10^4$/0.17 & \textbf{13/17/0.14}/950/\textbf{0.090}
    \\

    \bottomrule
  \end{tabular}
  }
\end{table*}